\def\year{2022}\relax
\documentclass[letterpaper]{article} 
\usepackage{aaai22}  
\usepackage{times}  
\usepackage{helvet}  
\usepackage{courier}  
\usepackage[hyphens]{url}  
\usepackage{graphicx} 
\usepackage{natbib}  
\usepackage{caption} 
\DeclareCaptionStyle{ruled}{labelfont=normalfont,labelsep=colon,strut=off} 
\usepackage{algorithm}
\usepackage{algorithmic}

\usepackage{newfloat}
\usepackage{listings}
\floatstyle{ruled}
\newfloat{listing}{tb}{lst}{}
\floatname{listing}{Listing}

\graphicspath{{./}{../}}

\newcommand{\minisection}[1]{\vspace{0.02in} \noindent {\bf #1}\ }
\usepackage{amsmath}
\usepackage{amssymb}
\usepackage{multirow}
\usepackage{subcaption}
\usepackage[final]{pdfpages}

\usepackage{xcolor}

\title{Incremental Meta-Learning via Episodic Replay Distillation \\ for Few-Shot Image Recognition}
\author {
    Kai Wang,\textsuperscript{\rm 1}
    Xialei Liu, \textsuperscript{\rm 2}
    Andy Bagdanov, \textsuperscript{\rm 3}
    Luis Herranz, \textsuperscript{\rm 1}
    Shangling Jui, \textsuperscript{\rm 4}
    Joost van de Weijer \textsuperscript{\rm 1}
}
\affiliations {
    \textsuperscript{\rm 1}  Computer Vision Center, Universitat Autònoma de Barcelona, Barcelona, Spain\\
    \textsuperscript{\rm 2}  College of Computer Science, Nankai University, Tianjin, China\\
    \textsuperscript{\rm 3} MICC, University of Florence, Florence, Italy\\
    \textsuperscript{\rm 4} Huawei Kirin Solution, Shanghai, China\\
    \{kwang, lherranz, joost\}@cvc.uab.es, xialei@nankai.edu.cn, andrew.bagdanov@unifi.it, jui.shangling@huawei.com
}

\begin{document}


\maketitle

\begin{abstract}
  Most meta-learning approaches assume the existence of a very large
  set of labeled data available for episodic meta-learning of base
  knowledge. This contrasts with the more realistic continual learning
  paradigm in which data arrives incrementally in the form of tasks
  containing disjoint classes. In this paper we consider this problem
  of Incremental Meta-Learning (IML) in which classes are presented
  incrementally in discrete tasks. We propose an approach to IML,
  which we call Episodic Replay Distillation (ERD), that mixes classes
  from the current task with class exemplars from previous tasks when
  sampling episodes for meta-learning. These episodes are then used
  for knowledge distillation to minimize catastrophic forgetting.
  Experiments on four datasets demonstrate that ERD surpasses the
  state-of-the-art. In particular, on the more challenging one-shot,
  long task sequence incremental meta-learning scenarios, we reduce
  the gap between IML and the joint-training upper bound from 3.5\% /
  10.1\% / 13.4\% with the current state-of-the-art to 2.6\% / 2.9\% /
  5.0\% with our method on Tiered-ImageNet / Mini-ImageNet / CIFAR100,
  respectively.
\end{abstract}

\section{Introduction} \label{sec:introduction}

Meta-learning, also commonly referred to as ``learning to learn'', is
a learning paradigm in which a model gains experience over a sequence
of learning episodes.\footnote{To avoid ambiguities, we use the term
  \textit{episode} in the sense used in meta-learning rather than how
  it is used in continual learning. We use \emph{task} in the sense of
  continual learning to refer to a disjoint group of new classes.}
This experience is optimized so as to improve the model's future
learning performance on unseen
tasks~\cite{hospedales2020meta}. Meta-learning is one of the most
promising techniques to learning models that can flexibly generalize,
like humans, to new tasks and environments not seen during
training. This capability is generally considered to be crucial for
future AI systems. Few-shot learning has emerged as the
paradigm-of-choice to test and evaluate meta-learning algorithms. It
aims to learn from very limited numbers of samples (as few as just
one), and meta-learning applied to few-shot image recognition in
particular has attracted increased attention in recent
years~\cite{su2020does,bateni2020improved,li2020adversarial,yang2021free}.

\begin{figure}[t]
\centerline{\includegraphics[width=0.9\linewidth]{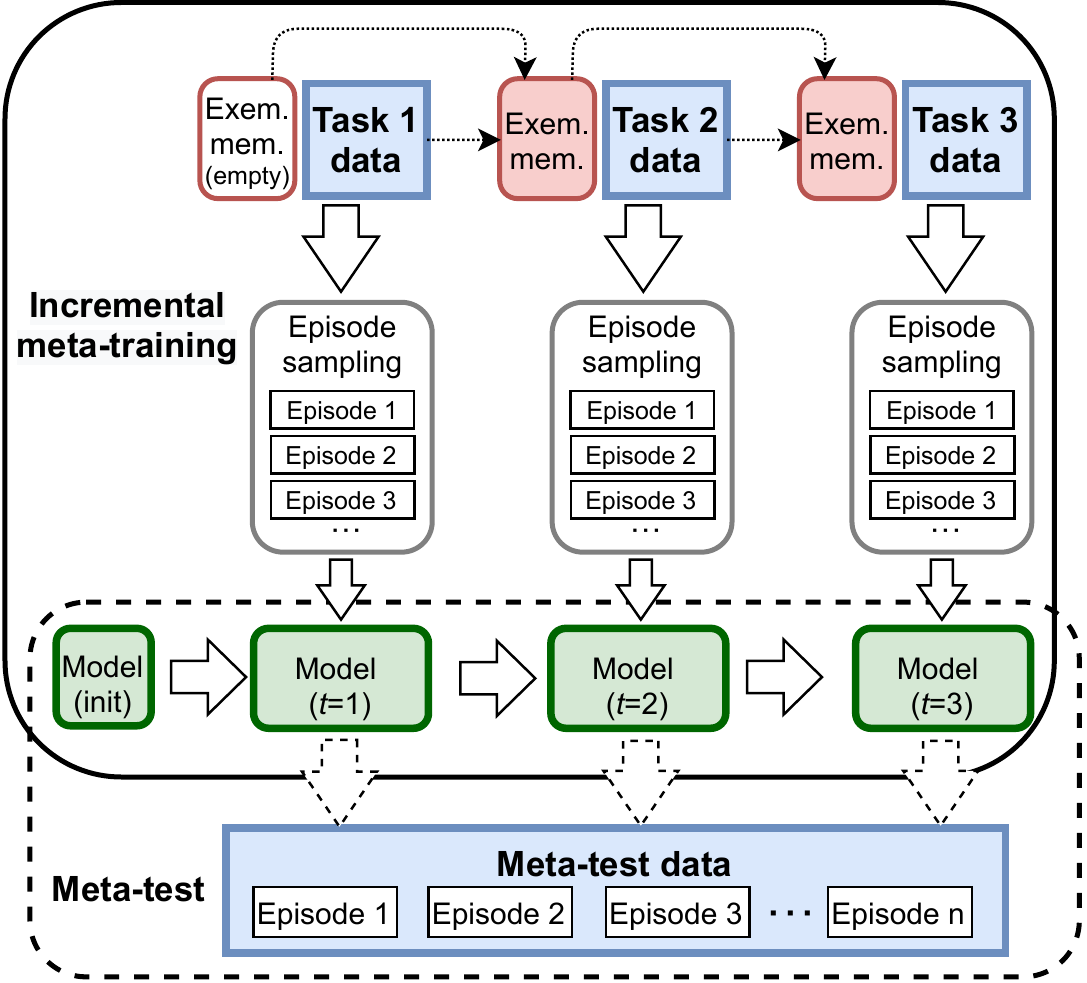}}
\caption{Incremental Meta-Learning (IML) with optional exemplar
  memories. Data from the previous tasks, unless in exemplar buffer,
  is unavailable in successive ones. Conventional meta-learning
  assumes a large number of base classes available for episodic
  training, while \emph{Incremental} Meta-Learning requires the
  meta-learner to update incrementally when a new set of classes (a new
  \emph{task}) arrives.}
\label{fig:iml_setting}
\end{figure}

However, most few-shot learning methods are limited in their learning
modes: they must train with a large number of classes, with a large
number of samples per class, and then to generalize and recognize new
classes from few samples. This can lead to poor performance in
practical incremental learning situations where the training tasks
arrive continually and there are insufficient categories at any given
time to learn a performant and general meta-model. The study of
learning from data that arrives in such a sequential manner is called
\emph{continual} or \emph{incremental}
learning~\cite{de2019continual,schwarz2018progress}. Catastrophic
forgetting is the main challenge of incremental learning
systems~\cite{mccloskey1989catastrophic}.
To address both the challenges of incremental and meta learning,
Incremental Meta-Learning (IML, illustrated
in Fig.~\ref{fig:iml_setting}) was recently proposed as a way of
applying few-shot learning in such incremental learning
scenarios~\cite{liu2020incremental}.

To address the IML problem, \citet{liu2020incremental} propose
the Indirect Discriminant Alignment (IDA) method. In this method,
class centers from previous tasks are represented by \emph{anchors} which are used to align
(by means of a distillation process) the old and new discriminants. They show that
this greatly reduces forgetting for short sequences of
tasks. They also extended IDA with exemplars\footnote{{\emph{Exemplars} refer
    to a small buffer of samples from previous tasks that can be used
    during the training of new ones. Note that the rest of samples are discarded and can not be accessed anymore.}} (EIML) from old
tasks, but surprisingly results showed that this fails to outperform
IDA without exemplars. This seems counter-intuitive, since exemplars
usually boost performance in incremental learning. We identify the
following drawbacks of IDA and EIML: (i) in IDA the anchors are fixed
after obtaining them from their corresponding tasks, while semantic
drift will gradually make prediction worse with successive tasks; (ii) in
EIML exemplars are used only for distillation and computing class
anchors, while they are not mixed with current tasks to make the
training more robust; and (iii) evaluation is only performed on short
sequences (maximally 3 tasks).

In this paper we propose Episodic Replay Distillation (ERD) to better
exploit saved exemplars and achieve significant improvement in
IML. ERD first divides episode construction into two parts: the
\emph{exemplar sub-episodes} containing only exemplars from past
tasks, and the \emph{cross-task sub-episodes} containing a mixture of
previous task exemplars and current task data. Exemplar sub-episodes
are then used to produce episode-level classifiers for distillation
over the query set. Cross-task sub-episodes combine previous task
exemplars with current task samples using a sampling probability
$P$. Since the current task contains more samples, and thus higher
diversity, a lower $P$ makes better use of the previous and the
current samples -- an interesting departure from conventional
continual learning in which we would typically desire \emph{more}
replay from past tasks.

The main contributions of this paper are:
\begin{itemize}
\item \textbf{Cross-task episodic training}: with our proposed
  \textit{cross-task sub-episodes}, in which the previous task
  exemplars and current task samples are mixed to work together to
  update the meta-learner.
\item \textbf{Episodic replay distillation}: different from EIML, our episodic
  replay distillation is on the episode level and not on class anchors, which in
  EIML are recomputed with all saved class exemplars.
\item \textbf{State-of-the-art results on all benchmarks}: ERD outperforms the
  state-of-the-art using both Prototypical Networks and Relation Networks. We
  are also the first to evaluate on long sequences of incremental
  meta-tasks.
\end{itemize}

\section{Related work} \label{sec:related_work}

\subsection{Few-shot learning} \label{subsec:few-shot}

Few-shot learning can be categorized into three main classes of
approaches: data augmentation, model enhancement and algorithm-based
methods~\cite{wang2020generalizing}. Among them, few-shot learning
based on metrics or optimization-based approaches are the main directions
of current research.

\minisection{Metric-based methods.} These approaches use embeddings learned from
other tasks as prior knowledge to constrain the hypothesis space. Since samples
are projected into an embedding subspace, the similar and dissimilar samples can
be easily discriminated. Among these techniques, ProtoNets~\cite{snell2017prototypical} and RelationNets~\cite{sung2018learning} are
the most popular.
    
\minisection{Optimization-based methods.} These use prior knowledge to search
for the model parameters which best approximate the hypothesis in search space
and use prior knowledge to alter the search strategy by providing good
initialization. Representative methods are
MAML~\cite{finn2017model} and Reptile~\cite{nichol2018first}.

\subsection{Continual learning} \label{subsec:continual_learning}
Continual learning methods can be divided into three main
categories~\cite{de2019continual}: replay-based, regularization-based
and parameter-isolation methods. We discuss the first two
categories since they are most relevant.

\minisection{Replay methods.} 
These prevent forgetting by incorporating data (real
or synthetic) from previous tasks.
There are two main strategies: exemplar
rehearsal~\cite{wu2019large,hou2019learning}, which
store a small number of training samples (called exemplars) from
previous tasks, and
pseudo-rehearsal~\cite{hayes2019remind,wu2018memory},
use generative models learned from previous task data distributions to
synthesize data. The classification model in continual learning is a
joint classifier and exemplars are used to correct the
bias or regularize the
gradients. However, in IML there is no joint
classifier (only a temporary classifier for each episode). Thus,
exemplar-based continual learning methods require adaptation to be
applicable to IML: replay for IML should be at the episode level
instead of the image level.

\minisection{Regularization-based methods.} These approaches add a
regularization term to the loss function which impedes changes to
parameters deemed relevant to previous
tasks~\cite{li2017learning,kirkpatrick2017overcoming}. Knowledge
distillation is popular in continual learning, which aims to either
align the outputs at the feature
level~\cite{liu2020generative} or the predicted
probabilities after a softmax layer~\cite{li2017learning}. However,
aligning at the feature level has been shown to be
ineffective~\cite{liu2020incremental} and the lack of a unified
classifier makes it impossible to align probabilities. Thus, we
propose to adapt knowledge distillation to the IML setup.

\subsection{Meta-learning for continual learning}
\label{subsec:meta_learning_for_cl}

In addition to the IML setting, there are a few continual learning
works that exploit meta-learning, such as
La-MAML~\cite{gupta2020maml} and
OSAKA~\cite{caccia2020online}. These methods focus on improving model
performance on task-agnostic incremental classification.  There are
also some works focusing on dynamic, few-shot visual recognition
systems, which aim to learn novel categories from only a few training
samples while at the same time not forgetting the base
categories~\cite{gidaris2018dynamic,yoon2020xtarnet}. Another related
setting is FSCIL~\cite{tao2020few}, where the authors constrain the
continual learning tasks using a few labeled samples.

Different from all these variants, the IML setting casts attention on
the original intention of few-shot learning: to make the model
generalize to unseen tasks even when training over incremental
tasks. Since the seen classes are increasing, the model should gain
more generalization ability instead of over-fitting to the current
task.

\section{Methodology} \label{sec:method}

In this section, we first define the standard few-shot learning
and introduce the \emph{incremental meta-learning}~(IML) setup. Then we describe our
approach to IML.

\subsection{Few-shot and meta-learning}
\label{subsec:few_shot_formula}
After introducing conventional few-shot learning, we then describe the incremental meta-learning approach.

\minisection{Conventional few-shot learning.} \label{subsubsec:few_shot_learning}
An approach to standard, non-incremental classification is to learn a parametric
approximation $p(y|x;\theta)$ of the posterior distribution of the class $y$
given the input $x$. Such models are trained by minimizing a loss function over
a dataset $D$ (e.g. the empirical risk). Few-shot learning, however, presents
extra difficulties since the number of samples available for each class $y$ is
very small (as few as one). In the meta-learning paradigm, training is divided
into two phases: meta-training, in which the model learns how to learn few-shot
recognition, and meta-testing where the meta-trained model is evaluated on
unseen few-shot recognition tasks.

Meta-training for few-shot learning consists of $T$ \emph{episodes}
(meta-training task in few-shot learning terminology), where each episode
$D^{\tau}$ is drawn from the train split of the entire dataset. Few-shot
recognition problems consisting of $N$ classes with $K$ training samples per
class are referred to $N$-way, $K$-shot recognition problems. Each episode is
divided into support set $S$ and query set $Q$: $D^{\tau} = (S, Q)$, where
$S = \{(x_{i}, y_{i})\}_{i=1}^{NK}$ consists of $N$ training classes each with
$K$ images, and $Q = \{(\hat x_i, \hat y_i)\}_{i=1}^{NK^Q}$ is a set of $K^Q$
images for each of the $N$ selected classes in the episode.

More specifically, we formulate our method based on
\textit{ProtoNets} in this section, and discuss its
extension to Relation Networks later. ProtoNets consist of an embedding
module $f_{\theta}$ and a classifier module $g$. First, the support set ${S}$ is
fed into the embedding module $f_{\theta}$ to obtain class prototypes
$\mathbf{c}_k$:
\begin{equation}
\label{eq:center}
    \mathbf{c}_k=\frac{1}{K} \!\! \sum_{(x_i,k) \in S} \!\! f_{\theta}(x_i).
\end{equation}
Then, an episode-specific classifier is applied to the query set, where the
prediction for class $k$ of query image $\hat x$ is:

\begin{equation}
\label{eq:protonet}
  g_{k}(f_{\theta}(S),f_{\theta}(\hat{x}))= p({y}=k|\hat{x};\theta)  = \frac{\exp(-d(f_\theta(\hat{x}),\mathbf{c}_k))}{\sum_{k'}
      \exp(-d(f_\theta(\hat{x}),\mathbf{c}_{k'}))}
\end{equation}
where the summation in the denominator is over all classes $k'$ in the support set. Then the meta-loss for updating $\theta$ is:
\begin{equation}
\begin{aligned}
\label{eq:ce_loss}
    L_{meta}(\theta;  S,Q)  = -
    \sum_{(\hat{x},\hat{y}) \in Q}[\log g_{\hat{y}}(f_{\theta}(S),f_{\theta}(\hat{x}))].
\end{aligned}
\end{equation}

\minisection{Incremental Meta-Learning (IML).} \label{subsubsec:iml}
When performing incremental meta-learning, data arrives as a sequence of disjoint
tasks: $T_{M}={X^{b}_{1}, ..., X^{b}_{t}, ..., X^{b}_{M}}$, where $M$ denotes
the number of tasks
, and $b$ the current training set
. The aim of IML is to incrementally learn the parameters $\theta_t$ for task $t$ from the disjoint tasks:
\begin{equation}
  \label{eq:iml_loss}
  \theta_t^{*}=\arg \min_{\theta_t} L(\theta_t;  \theta_{t-1},S_{t}, Q_{t}),
\end{equation}
Depending on whether we store exemplars from previous tasks, the support set
$S_{t}$ and query set $Q_t$ can be constructed differently using samples in the
current task and exemplars from previous tasks. These query and support sets are
described in detail in the next section.

\begin{figure}
\begin{center}
\includegraphics[width=0.94\linewidth]{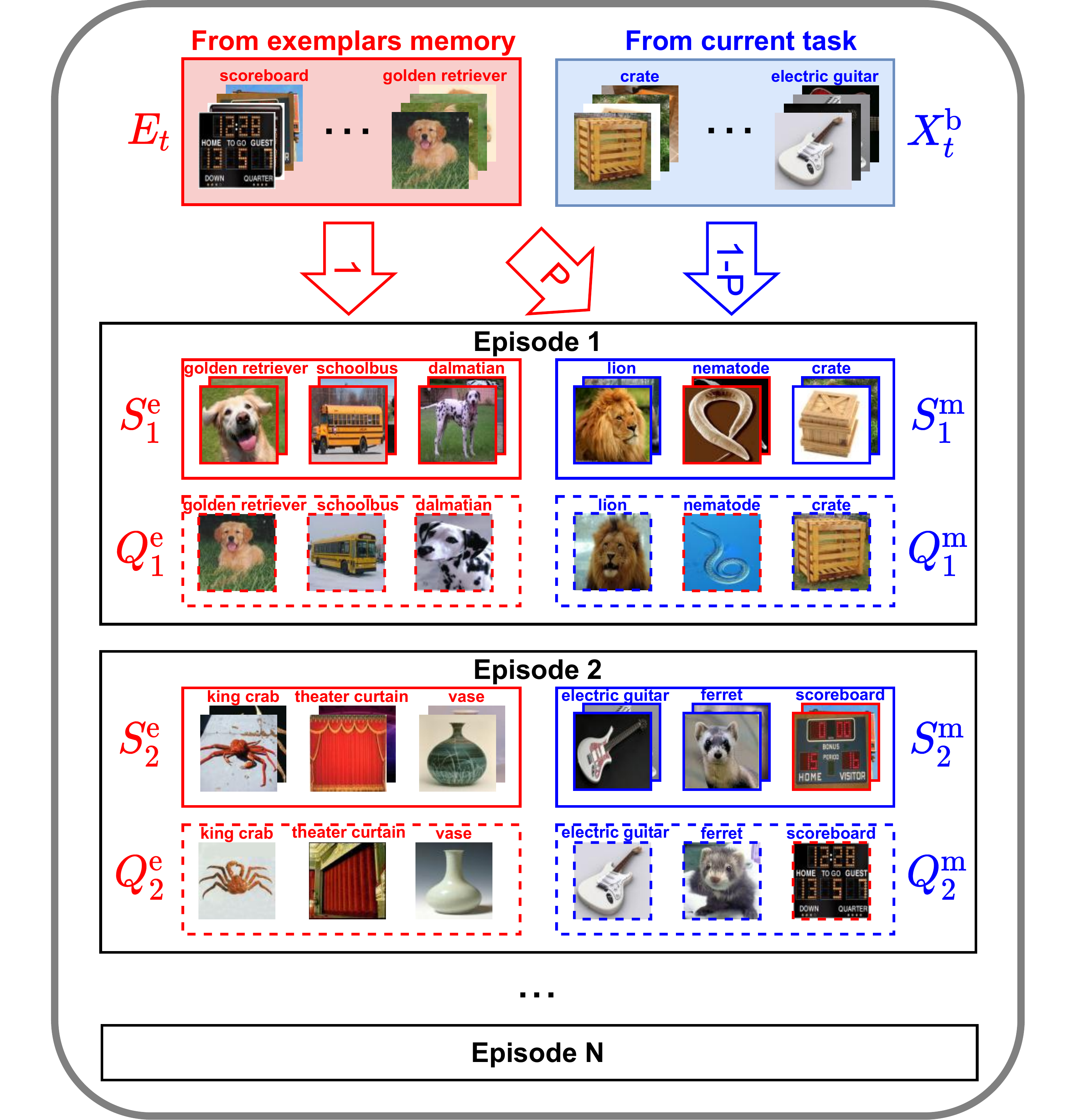}
\end{center}
\caption{Proposed episode sampling. During episodic meta-learning we
  build two sets of few-shot problems: \emph{exemplar sub-episodes} based on only exemplars from
  previous tasks ($S^e_i$ and $Q^e_i$) and \emph{cross-task sub-episodes} with a mix of exemplars from
  previous tasks and samples from the current task ($S^m_i$ and $Q^m_i$).}
\label{fig:episode_sampling}
\end{figure}

\subsection{Cross-task episodic training}
\label{subsec:crosstask}

Keeping exemplars from previous tasks is a successful approach to avoid catastrophic forgetting in conventional incremental learning. However,
it is not obvious how to leverage exemplars for incremental
\emph{meta}-learning. We propose a novel
way of using exemplars for this specific setup. 

We consider two strategies for exemplars. The first strategy stores $N_{ex}$ exemplars ($N_{ex}=20$ by default) for each class for each previous task, which is standard in replay-based continual learning methods (UCIR, PODNet, etc.). In this case, the buffer is linearly increased by training sessions. The second strategy fixes the maximum buffer size to $BF$ exemplars ($BF$=1000 by default). We apply both settings in the ablation study and will use the increasing buffer strategy as default.

To fully exploit exemplars from previous
tasks, for each episode during meta-learning we construct two sets of support
and query images (see Fig.~\ref{fig:episode_sampling}). Each episode is broken down into two sets of few-shot problems:

\begin{itemize}
\item In each episode we construct a \emph{cross-task sub-episode} by sampling
  $N$ classes from the current task with probability $1-P$ and from previous
  tasks with probability $P$. Thus, we have on average $N*(1-P)$ classes from
  the current task and $N*P$ from the past. Then, for each class we randomly
  sample $K$ images as support set $S^{m}$ and $K_{Q}$ images as query set
  $Q^{m}$ ($m$ denotes that we \textit{mix} the
  exemplars with current task samples here).
  
\item We also construct an \emph{exemplar sub-episode} by sampling $N$ classes
  from the only the exemplars from previous tasks, each with $K+K_{Q}$ images to
  form a support set $S^{e}$ and query set $Q^{e}$. Note that this episode is
  only composed of \textit{exemplars} from previous tasks.
\end{itemize}

The reason we sample cross-task sub-episodes with probability $P$ is that exemplars
are normally much fewer than samples in the current task, and thus the exemplars
are not expected to be as varied as the samples from the current task. With a
probability $P$, we can control the balance between current and previous classes in the cross-task sub-episode. And it doesn't influence the update of the memory buffer.

Given $S^{m},Q^{m}$, the meta-training loss is defined as:

\begin{equation}
\begin{aligned}
\label{eq:meta_loss}
    L_{meta}(\theta_t;  S^m,Q^m)  = 
    \sum_{(\hat{x},\hat{y}) \in Q^m}[-\log g_{\hat{y}}(f_{\theta_t}(S^m),f_{\theta_t}(\hat{x}))].
\end{aligned}
\end{equation}
This loss is only computed over $S^{m}$ and $Q^{m}$ since in $S^{m}$ we have
samples from the previous and current tasks.

\subsection{Episodic Replay Distillation (ERD)}
\label{subsec:erd_iml}

\begin{figure}
\begin{center}
\includegraphics[width=1.0\linewidth]{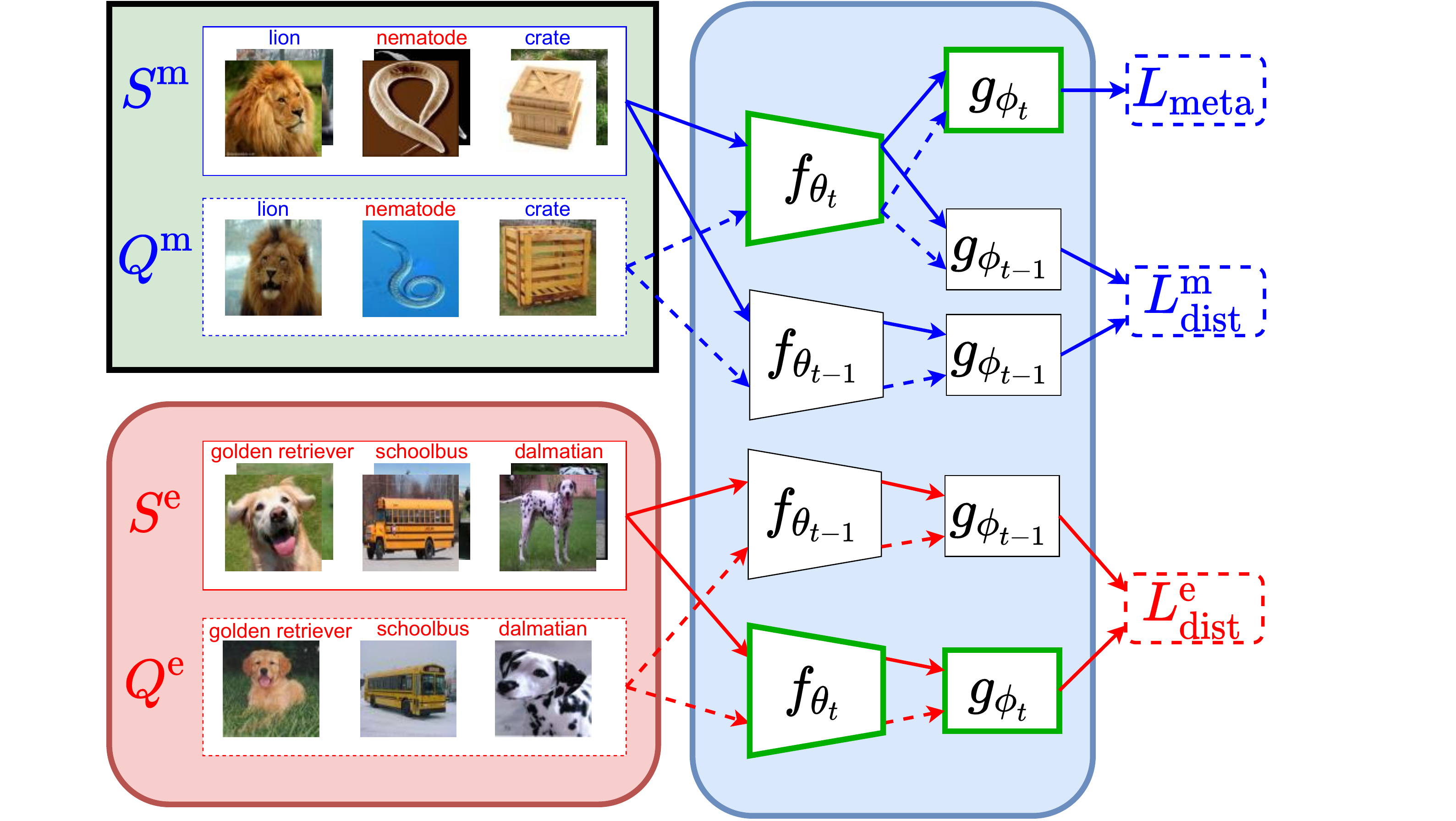}
\end{center}
\caption{Proposed Episodic Replay Distillation framework. Modules in green are
  the current embedding model, which are updated with both cross-task and exemplar sub-episodes. Red
  lines and blue lines are data flows for exemplar sub-episode and cross-task sub-episode,
  respectively. Solid lines and dotted lines indicate the data flows from
  support set and query set respectively. When computing loss for ProtoNets, $g$ is a parametric-free operation,
  while for Relation Networks, $g$ consists of a set of parameters $\phi$.}
\label{fig:relation_erd}
\end{figure}

In addition to cross-task episodic training, multiple distillation losses are
applied to avoid forgetting when we update the current model (see Fig.~\ref{fig:relation_erd}). We first explore
distillation using \textit{exemplar sub-episodes}. This is computed as:
\begin{equation}
\begin{aligned}
\label{eq:distill_exem_loss}
    L_{dist}^{e}  (\theta_{t};\theta_{t-1},S^{e},Q^{e}) = \sum_{ \hat{x} \in Q^{e}} \{ KL[ g(f_{\theta_{t-1}}(S^{e}),f_{\theta_{t-1}}(\hat{x})) \\ || \, g(f_{\theta_{t}}(S^{e}),f_{\theta_{t}}(\hat{x}))] \}
\end{aligned}
\end{equation}

where $f_{\theta_{t-1}}$ is the embedding network from the previous task with
parameters $\theta_{t-1}$. During training, only the current model
$f_{\theta_{t}}$ is updated and $f_{\theta_{t-1}}$ is frozen.

Next, similar to Eq.~\ref{eq:distill_exem_loss}, we also propose a
distillation loss using \textit{cross-task sub-episodes}. It
is computed according to:

\begin{equation}
\begin{aligned}
\label{eq:distill_base_loss}
    L_{dist}^{m}  (\theta_{t};\theta_{t-1},S^{m},Q^{m}) = \sum_{\hat{x} \in Q^{m}}  \{ KL[  g(f_{\theta_{t-1}}(S^{m}),f_{\theta_{t-1}}(\hat{x})) \\ || \, g(f_{\theta_{t}}(S^{m}),f_{\theta_{t}}(\hat{x}))] \}
\end{aligned}
\end{equation}
The only difference between this distillation loss function and
Eq.~\ref{eq:distill_exem_loss} is the inputs. Finally, $\theta_{t}$ is updated by minimizing:

\begin{equation}
\begin{aligned}
\label{eq:erd_loss}
L(\theta_t;  \theta_{t-1},S^{e},Q^{e},  S^{m},Q^{m}) & =L_{meta}   + \lambda_m  \cdot L_{dist}^{m}   + \lambda_e \cdot L_{dist}^{e}  
\end{aligned}
\end{equation}
$\lambda_m$ and $\lambda_e$ are trade-off parameters.

\subsection{Extension to Relation Networks} \label{subsec:extension_relation_nets}

Episodic Replay Distillation is not limited to ProtoNets. It can also be extended to Relation Networks~\cite{sung2018learning}, which consist of a relation module
with parameters $\phi$. Losses introduced in previous sections are adapted as:
\begin{equation}
\begin{aligned}
\label{eq:relation_meta_loss}
    L_{meta}  (\theta_t, \phi_t;  S^m,Q^m) & = \\
   \sum_{({x},{y})  \in S^m, (\hat{x},\hat{y}) \in Q^m}   & [ g_{\phi_{t}}(\mathcal{C}(f_{\theta_t}(x),f_{\theta_t}(\hat{x})))-\mathbf{1}(y=\hat{y})]^{2},
\end{aligned}
\end{equation}
where $\mathcal{C}$ is the concatenation of support set and query set
embeddings, $\mathbf{1}$ is the Boolean function returning 1 when its argument
is true and 0 otherwise. Distillation losses are updated as (MSE denotes mean squared error):
\begin{equation}
\begin{aligned}
\label{eq:relation_distill_base_loss}
    L_{dist}^{m}  (\theta_{t},\phi_t;\theta_{t-1},S^{m},Q^{m}) & =  \\ \sum_{{x} \in S^{m},\hat{x} \in Q^{m}}  \{ \mathrm{MSE} & [  g_{\phi_{t-1}}(\mathcal{C}(f_{\theta_{t-1}}(x),f_{\theta_{t-1}}(\hat{x}))),  \\ & g_{\phi_{t-1}}(\mathcal{C}(f_{\theta_{t}}(x),f_{\theta_{t}}(\hat{x})))] \} \\
    L_{dist}^{e}  (\theta_{t},\phi_t;\theta_{t-1},\phi_{t-1},S^{e},Q^{e}) & =  \\ \sum_{{x} \in S^{e},\hat{x} \in Q^{e}}\{ \mathrm{MSE}  [ &  g_{\phi_{t-1}}   (\mathcal{C}(f_{\theta_{t-1}}(x),f_{\theta_{t-1}}(\hat{x}))),  \\ & g_{\phi_{t}}  ( \mathcal{C}(f_{\theta_{t}}(x),f_{\theta_{t}}(\hat{x})))] \},
\end{aligned}
\end{equation}

Although Relation Networks and ProtoNets adopt different ways to calculate the
prediction probabilities for given query images, they share similar network
architectures with embedding and classification modules. This type of
architecture is widely used in metric-based few-shot learning and we believe
that our method can be easily adapted to other methods with similar
architectures.

\section{Experiments} \label{sec:experiments}

In this section we report on a range of experiments to quantify the contribution
of each element of our proposed approach and to compare our performance against
the state-of-the-art in continual few-shot image classification.

\subsection{Experimental setup} \label{subsec:experimental_setup}

Here we describe the datasets and experimental protocols in our
experiments. Source code is in supplementary material.

\begin{table}[htbp!]
\begin{center}
\scalebox{0.77}{
\begin{tabular}{|r|c|c|c|c||c|}
\hline
& \multicolumn{4}{|c||}{\textit{IML training tasks}} &\multirow{2}{*}{Meta-test Images} \\
\cline{0-4}
Task \#: & 1 & 2 & ... & 16 & \\
\hline

Classes per task: & 5 & 5 &\multirow{3}{*}{...} & 5 & 20 \\
\cline{0-2}
\cline{5-6}

Images in train split: & 500 & 500 &  & 500 & \multirow{2}{*}{600} \\

\cline{0-2}
\cline{5-5}

Images in test split: & 100 & 100 &  & 100& \\

\hline
\end{tabular}
}
\end{center}
\caption{Our proposed 16-task split of  Mini-ImageNet and CIFAR100 datasets for incremental few-shot learning.
}
\label{tab:splitting}
\end{table}

\minisection{Datasets.} We evaluate performance on four datasets:
Mini-ImageNet~\cite{vinyals2016matching}, CIFAR100~\cite{krizhevsky2009learning}, CUB-200-2011~\cite{WahCUB_200_2011} and Tiered-ImageNet~\cite{ren2018meta}. Mini-ImageNet consists of 600
84$\times$84 images from 100 classes. We propose a split with 20 of these
classes as \textit{meta-test set} unseen in all training sessions. The other 80
classes are used to form the incremental meta-training set which is split into 4
or 16 tasks with equal numbers of classes for incremental meta-learning. Each
class in each task is then divided into a meta-training split with 500 images,
from which support and query sets are sampled for each episode, and a test split
with 100 images that is set aside for task-specific evaluation. We select
$N_{ex}=20$ exemplars per class before proceeding to the next task. Table~\ref{tab:splitting} is an illustration of the 16-task setting data split.

CIFAR100 also contains 100 classes, each with 600 images, so we use the same
splitting criteria as for Mini-ImageNet. The CUB dataset contains 11,788 images
of 200 birds species. We split 160 classes into an incremental meta-training set
and the other 40 are kept as a meta-test set of unseen classes. We divide the
160 classes into 4 or 16 equal incremental meta-learning tasks. Since there are fewer
images per class, we choose $N_{ex}=10$ images per class as exemplars for each
previous task and 20 images as test split for each task. On Tiered-ImageNet, We keep the same test split (8 categories, 160 classes) as in the original setup, then split the training and validation classes (26 categories, 448 classes) into 16 equal tasks. And since in this case, each task is with more classes than other datasets setup, it's a much easier setting, we only test it under the 16-task setup.

\minisection{Implementation details.} We use
ProtoNets as our main meta-learner, but also verify
ERD using Relation Networks. We evaluate both the widely
used 4-Conv~\cite{snell2017prototypical} and
ResNet-12~\cite{he2016deep} as feature extractors. We sample 200 and 50 episodes
for 4-task and 16-task setups per task in each training epoch respectively. We train each meta-learning task for 200 epochs by
Adam.

We evaluate on two widely used few-shot setups: 1-shot/5-way and
5-shot/5-way. We include results on both incremental training tasks
and the unseen \textit{meta-test set}. For each task (including the
unseen set), we randomly construct $N_{ep}$ episodes to obtain the final performance
of the meta-learner, which is computed as the mean and 95\% confidence interval
of the classification accuracy across $N_{ep}$ episodes. $N_{ep}$ is 10000 for the 4-task and 1000 for the 16-task setup. Here we only report the
mean accuracy, but confidence intervals are reported in the supplementary
material. For exemplar selection for ProtoNets, we use the Nearest-To-Center (NTC) criterion to select
the samples closest to the class-mean. For Relation
Networks, since the image embeddings are feature maps
instead of feature vectors, we cannot obtain the class prototypes and therefore
use random selection. By default, we set $\lambda_m=\lambda_e=0.5, P=0.2$.

\minisection{Compared methods.} We compare our method with a finetuning baseline
(FT), IDA~\cite{liu2020incremental}, and a variant of IDA with $N_{ex}$ exemplars per class
(EIML). The meta-test upper bounds are obtained by
jointly training on all training tasks and testing on the unseen meta-test
split (i.e. the standard setting in non-incremental few-shot learning). We evaluate on two sets of tasks separately for comparison. At each
training session, we evaluate on previously \textit{seen} classes as way of
measuring forgetting. This we call \textit{mean accuracy on seen classes}.
Performance on the \textit{meta-test set} (all unseen classes) is to show the
generalization ability.

\subsection{Experiments results}
\begin{figure*}[htbp!]

\begin{minipage}[b]{0.245\linewidth}
\centering
\includegraphics[width=\textwidth]{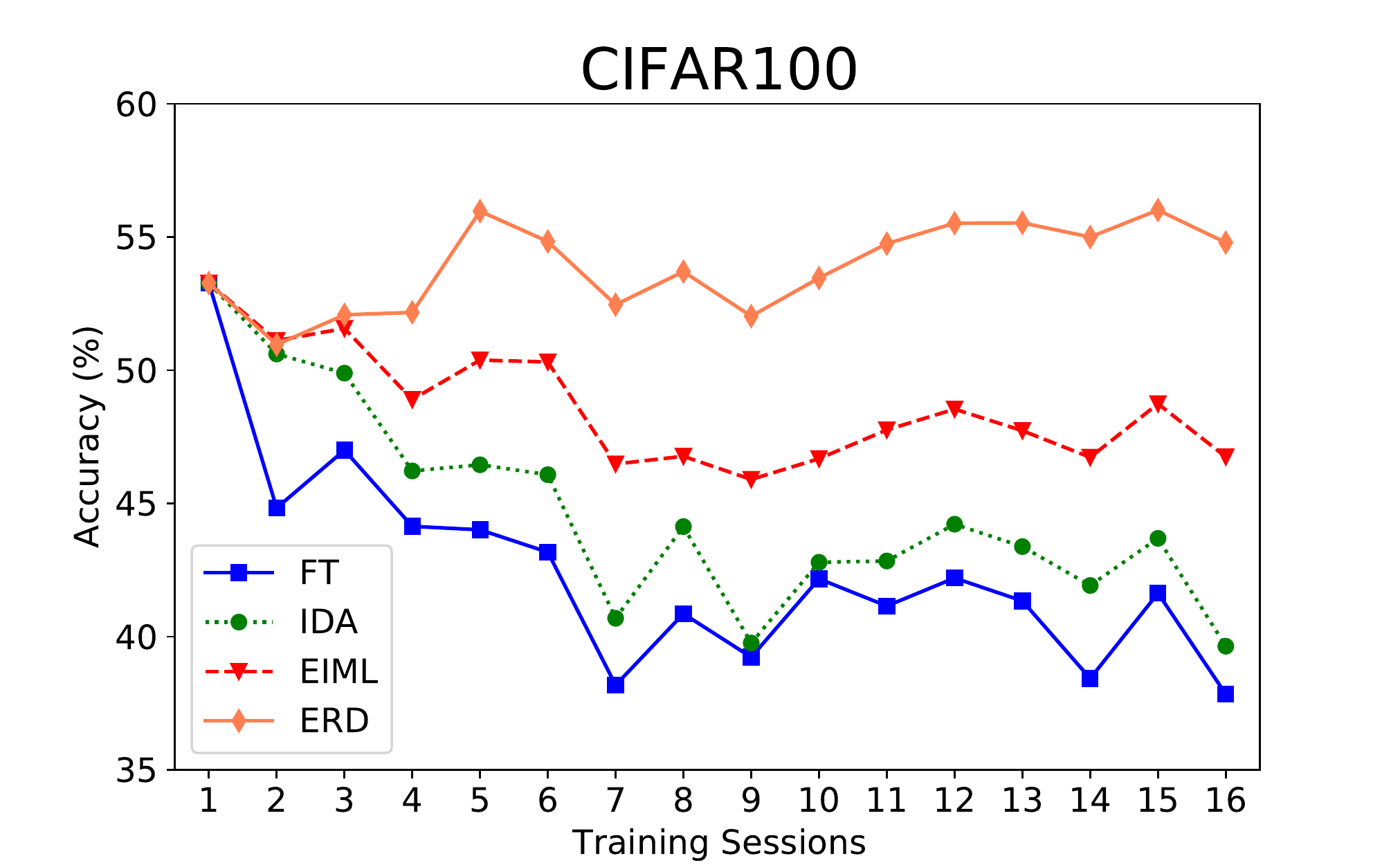}
\subcaption{1-shot mean accuracy on seen}
\label{fig:1shot_16task_cifar100_seen}
\end{minipage}
\begin{minipage}[b]{0.245\linewidth}
\centering
\includegraphics[width=\textwidth]{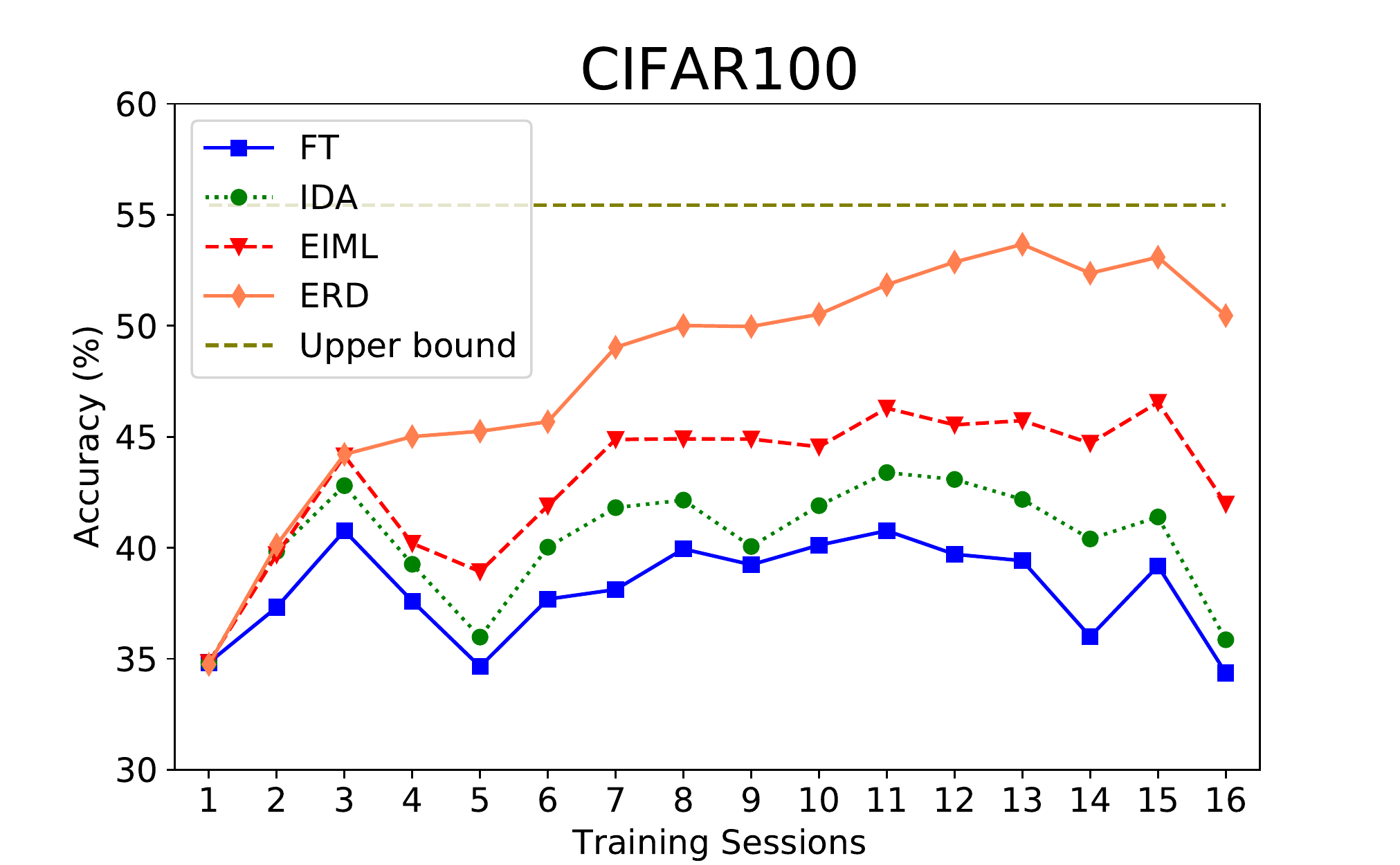}
\subcaption{1-shot meta-test accuracy}
\label{fig:1shot_16task_cifar100_meta}
\end{minipage}
\begin{minipage}[b]{0.245\linewidth}
\centering
\includegraphics[width=\textwidth]{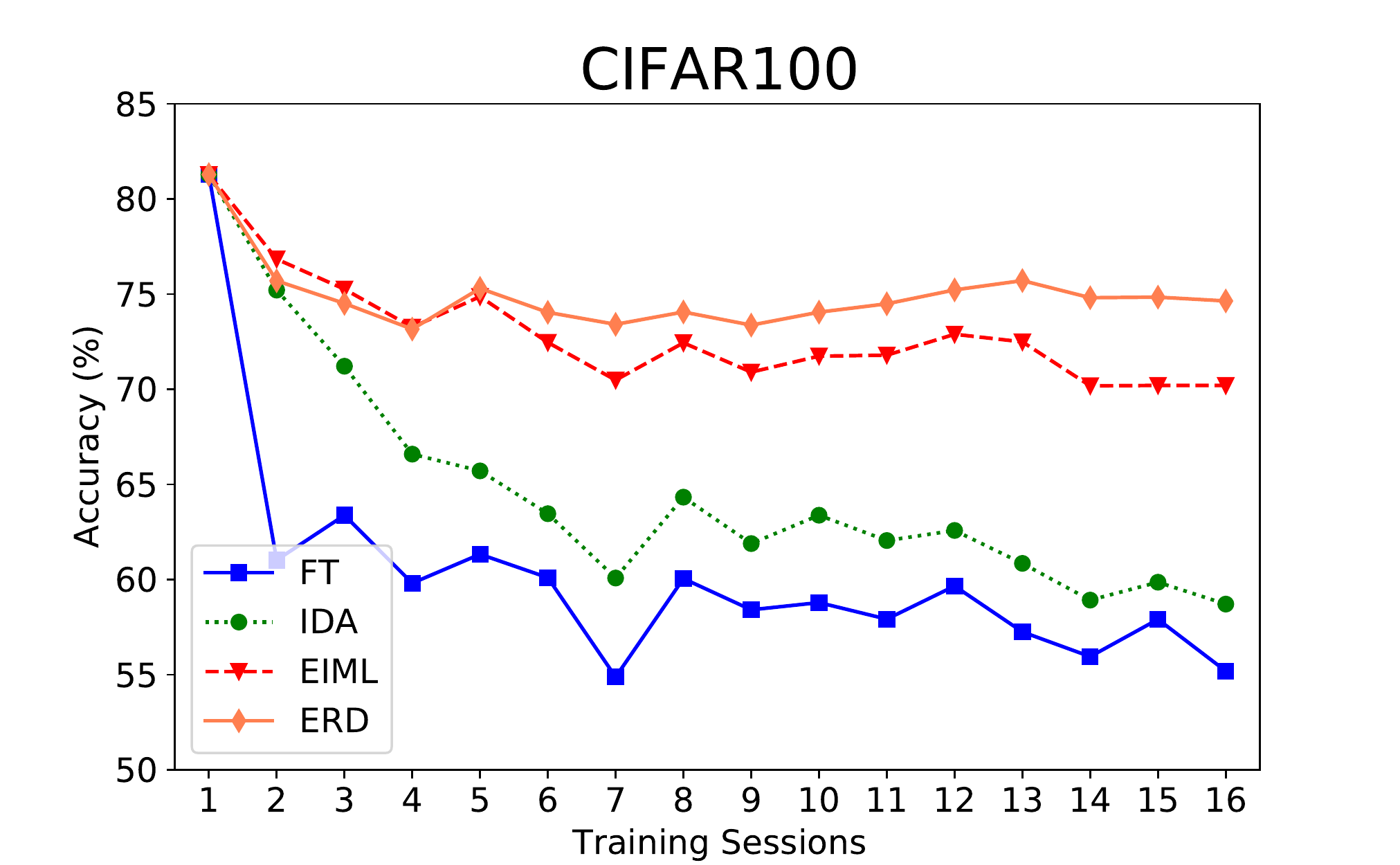}
\subcaption{5-shot mean accuracy on seen}
\label{fig:5shot_16task_cifar100_seen}
\end{minipage}
\begin{minipage}[b]{0.245\linewidth}
\centering
\includegraphics[width=\textwidth]{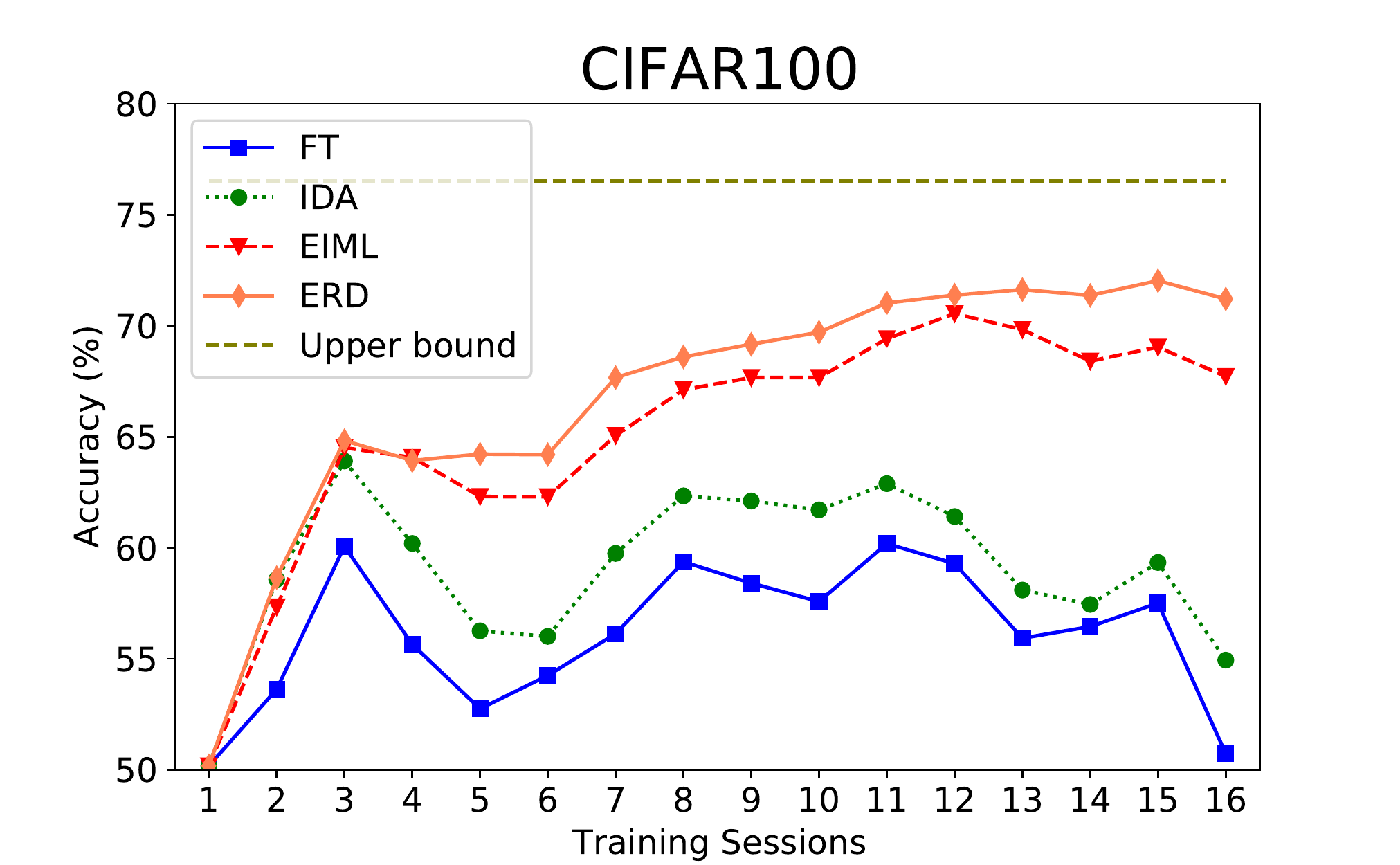}
\subcaption{5-shot meta-test accuracy}
\label{fig:5shot_16task_cifar100_meta}
\end{minipage}

\begin{minipage}[b]{0.245\linewidth}
\centering
\includegraphics[width=\textwidth]{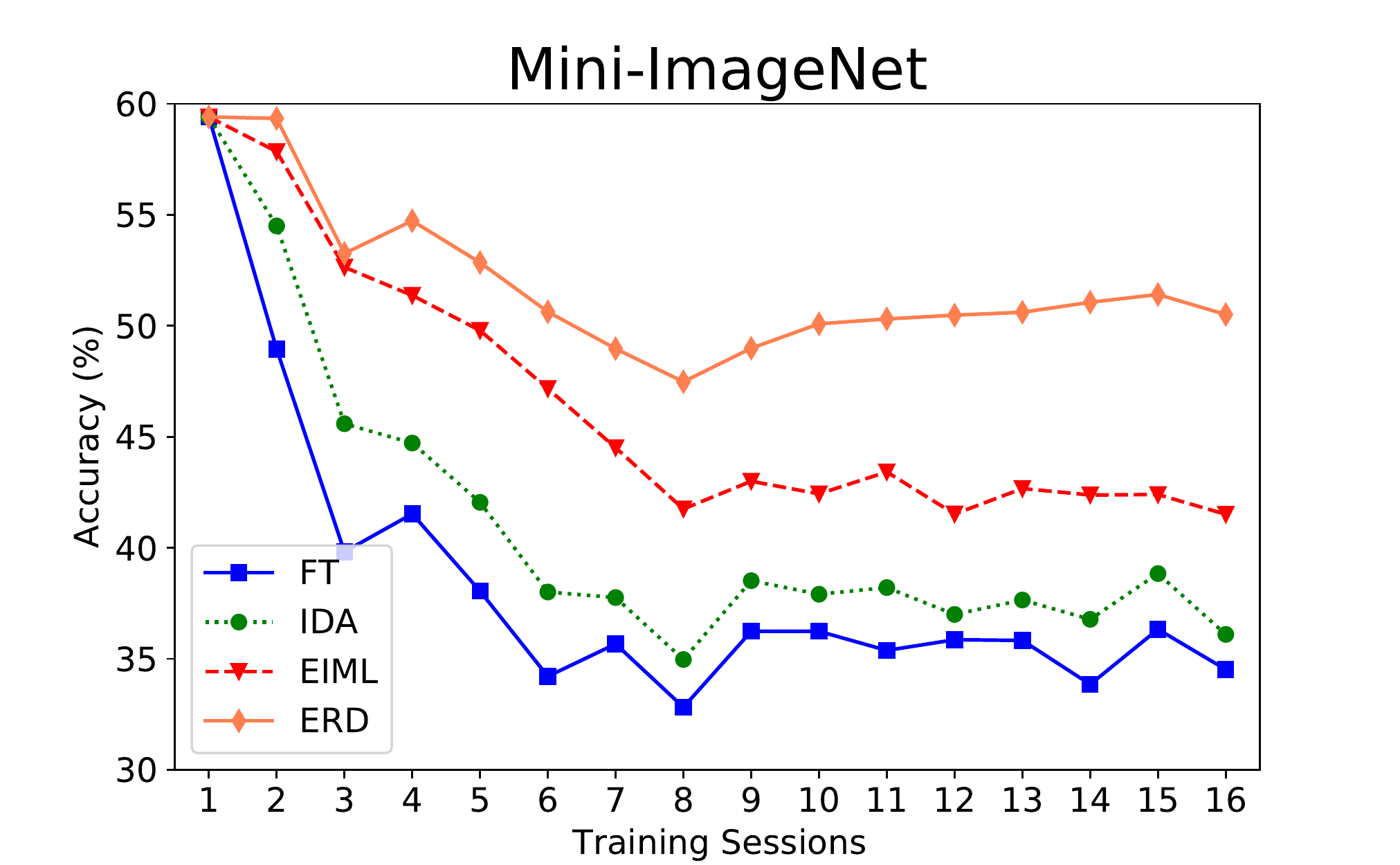}
\subcaption{1-shot mean accuracy on seen}
\label{fig:1shot_16task_mini_seen}
\end{minipage}
\begin{minipage}[b]{0.245\linewidth}
\centering
\includegraphics[width=\textwidth]{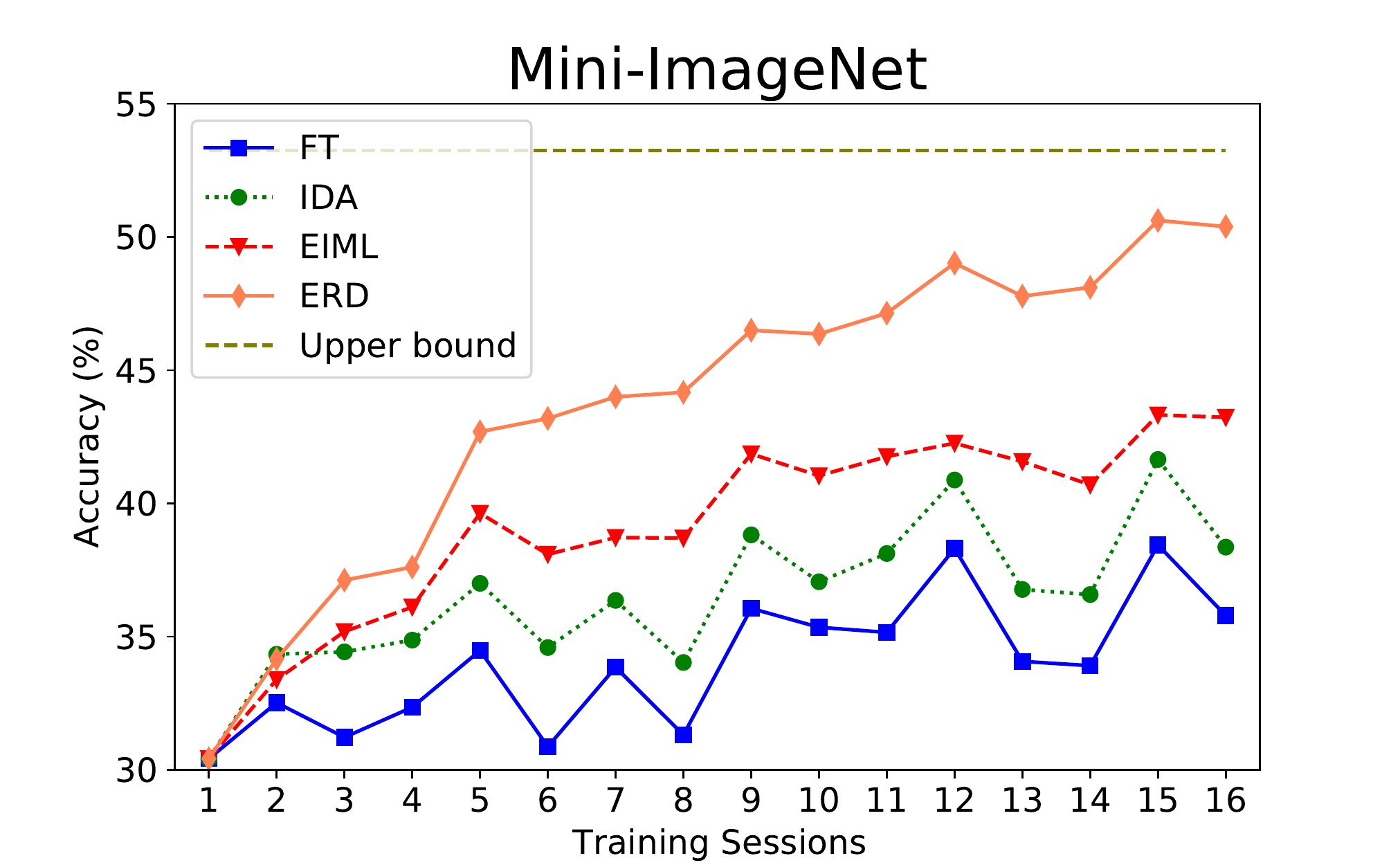}
\subcaption{1-shot meta-test accuracy}
\label{fig:1shot_16task_mini_meta}
\end{minipage}
\begin{minipage}[b]{0.245\linewidth}
\centering
\includegraphics[width=\textwidth]{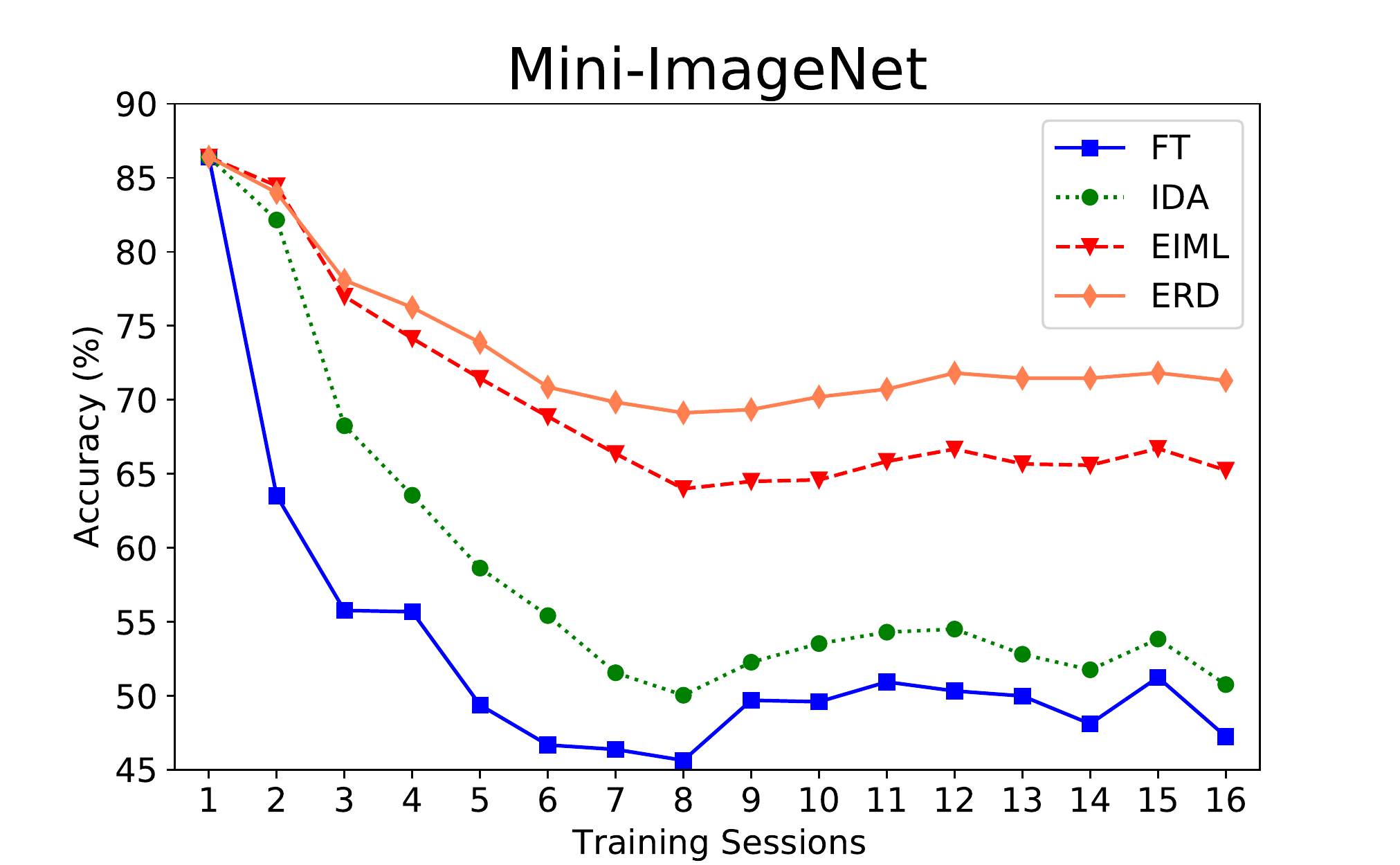}
\subcaption{5-shot mean accuracy on seen}
\label{fig:5shot_16task_mini_seen}
\end{minipage}
\begin{minipage}[b]{0.245\linewidth}
\centering
\includegraphics[width=\textwidth]{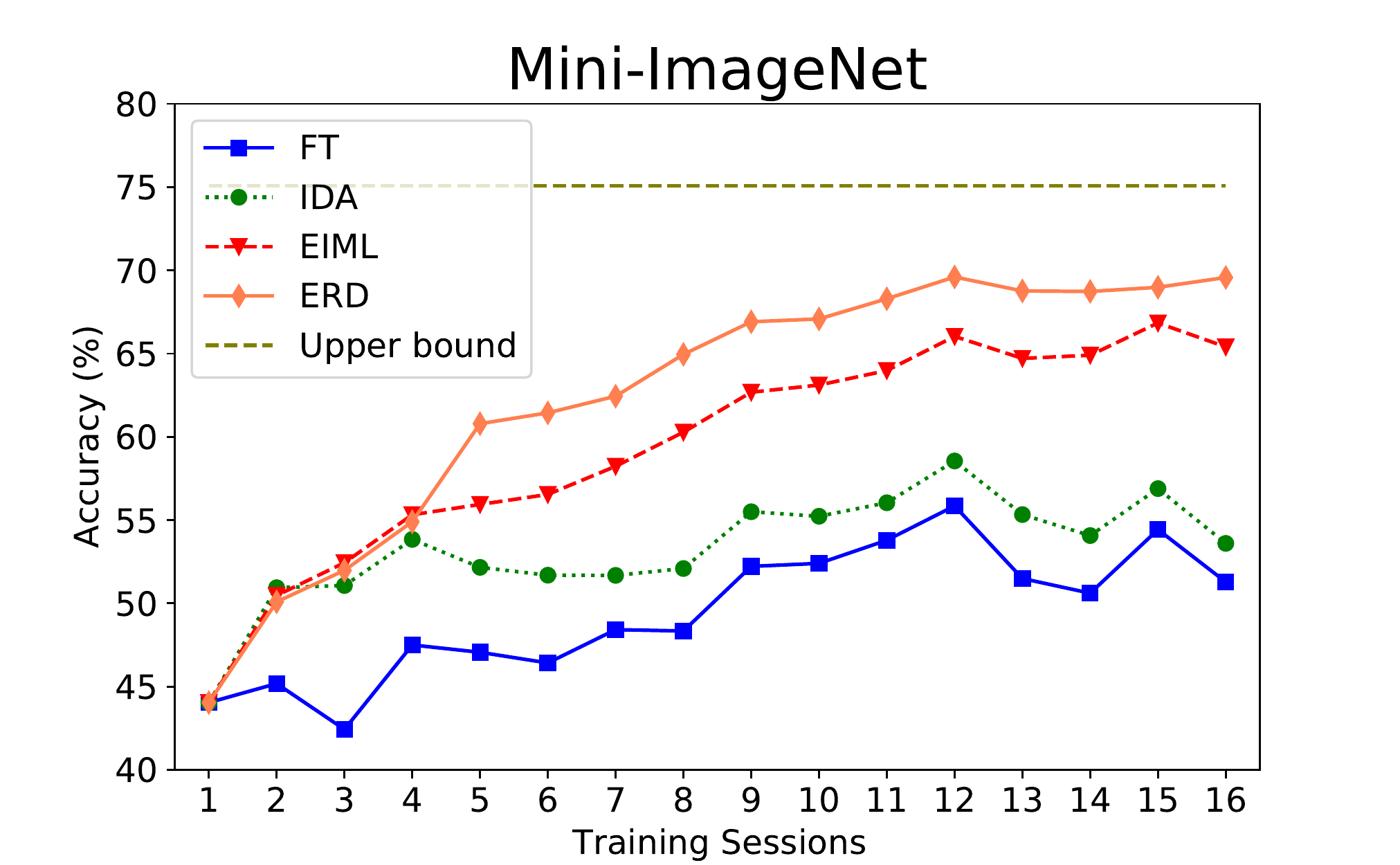}
\subcaption{5-shot meta-test accuracy}
\label{fig:5shot_16task_mini_meta}
\end{minipage}

\begin{minipage}[b]{0.245\linewidth}
\centering
\includegraphics[width=\textwidth]{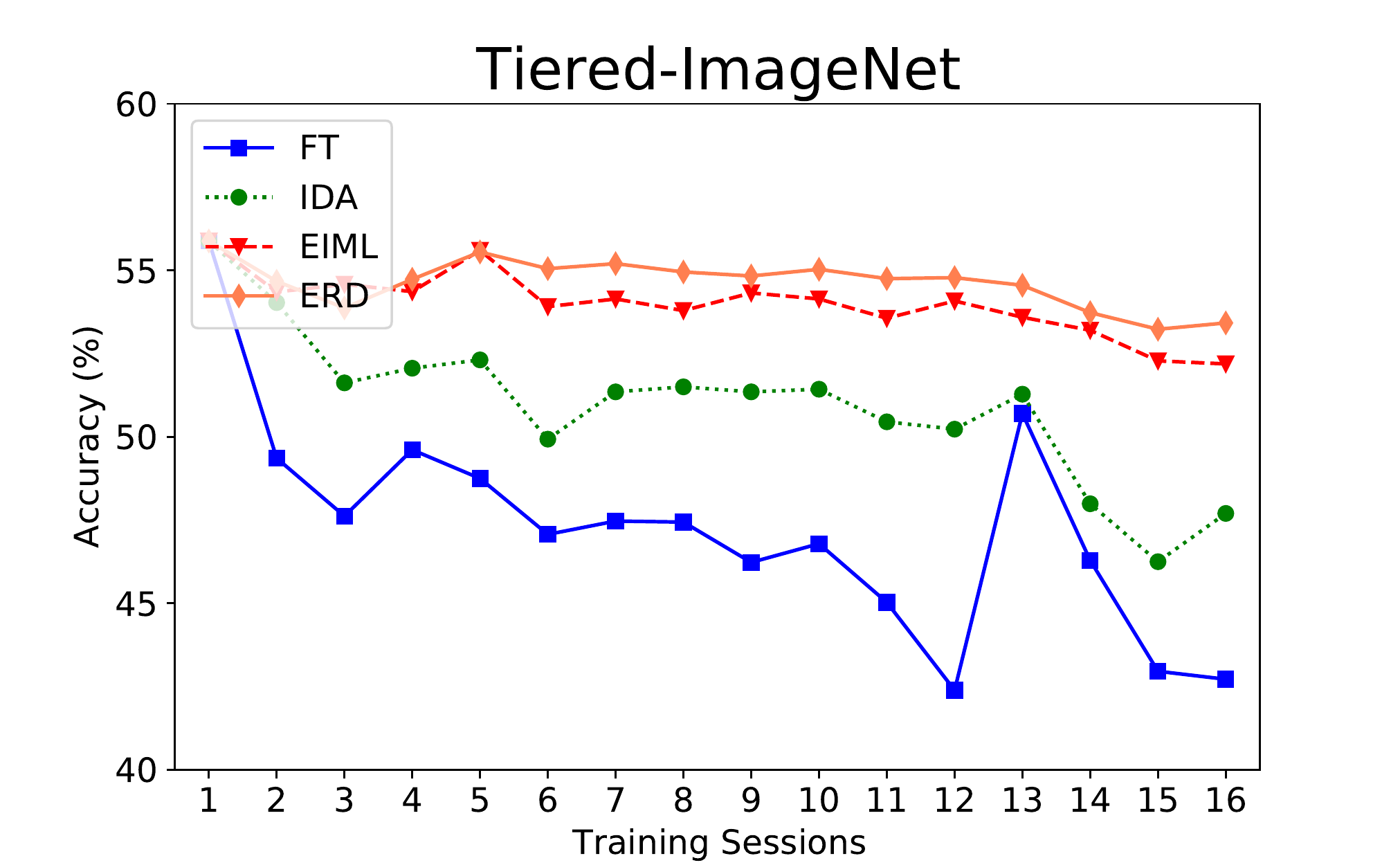}
\subcaption{1-shot mean accuracy on seen}
\label{fig:1shot_16task_tiered_seen}
\end{minipage}
\begin{minipage}[b]{0.245\linewidth}
\centering
\includegraphics[width=\textwidth]{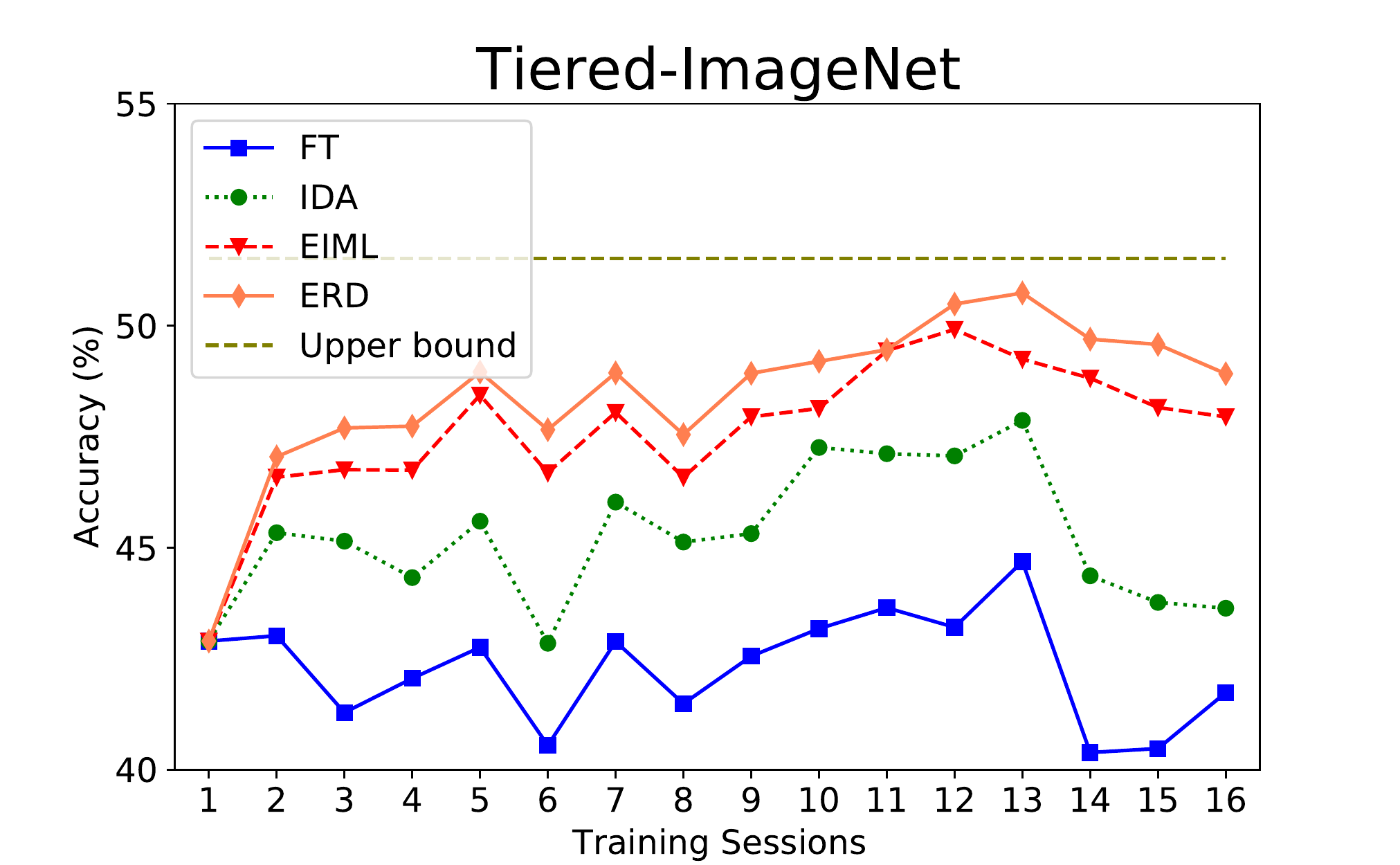}
\subcaption{1-shot meta-test accuracy}
\label{fig:1shot_16task_tiered_meta}
\end{minipage}
\begin{minipage}[b]{0.245\linewidth}
\centering
\includegraphics[width=\textwidth]{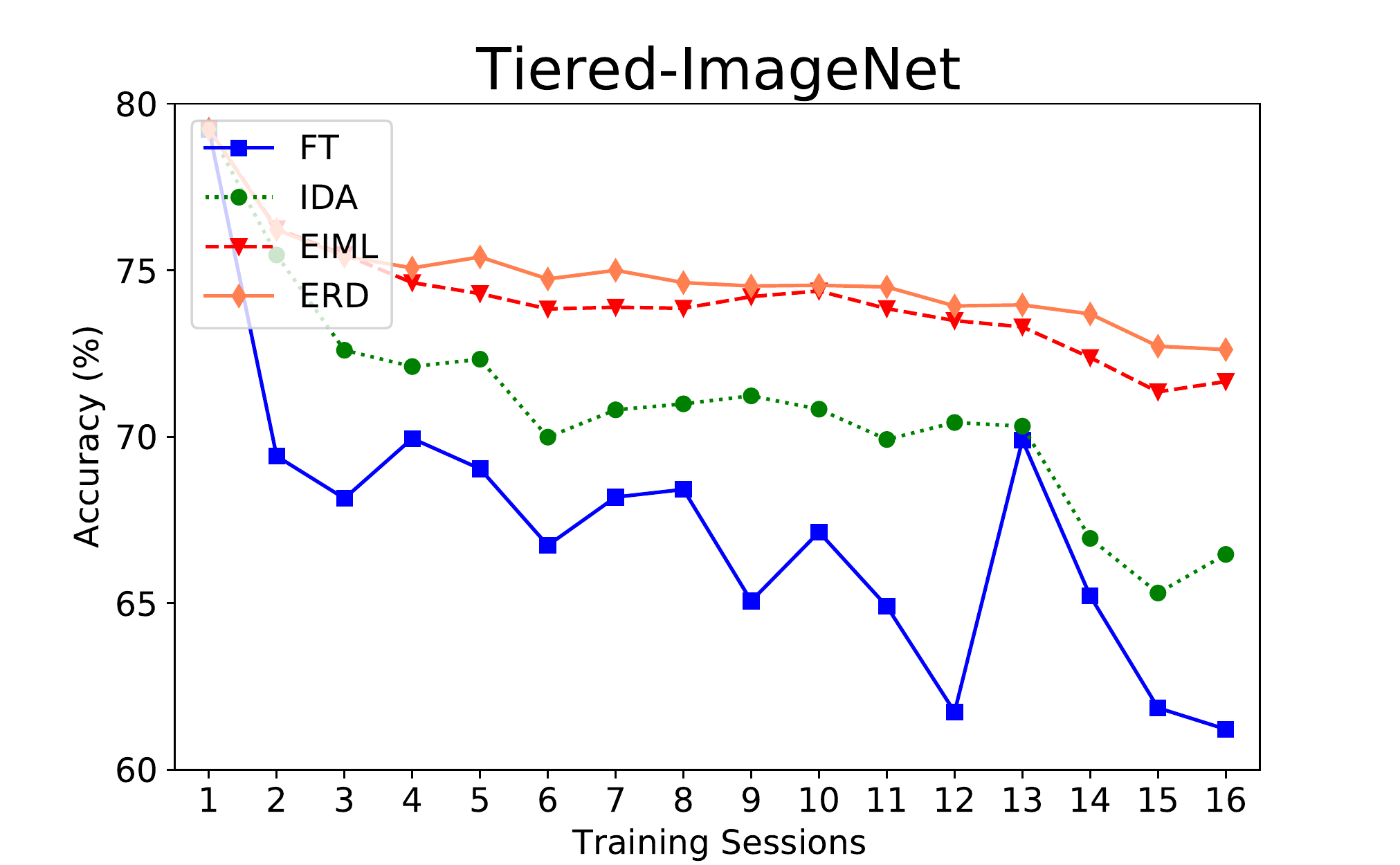}
\subcaption{5-shot mean accuracy on seen}
\label{fig:5shot_16task_tiered_seen}
\end{minipage}
\begin{minipage}[b]{0.245\linewidth}
\centering
\includegraphics[width=\textwidth]{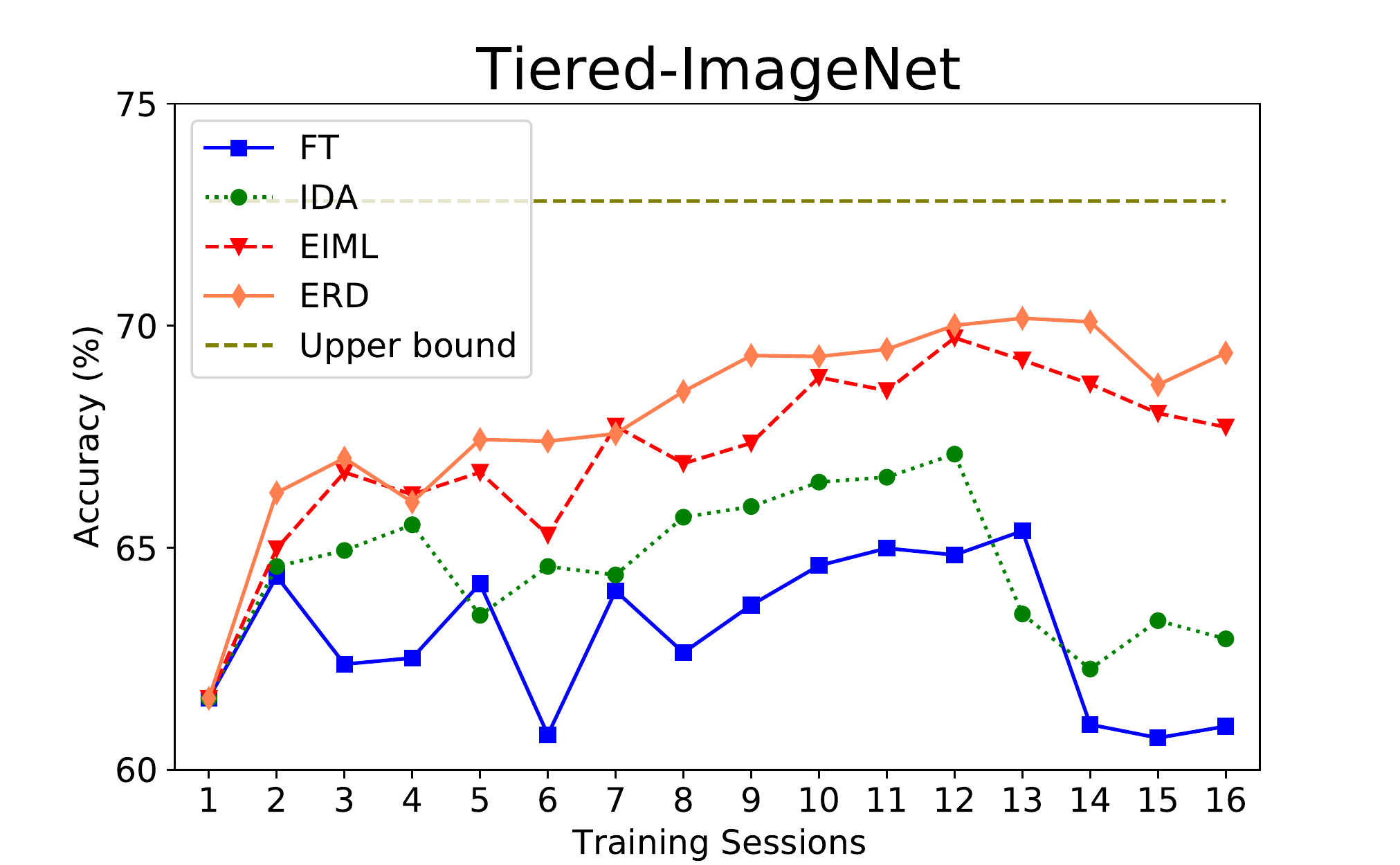}
\subcaption{5-shot meta-test accuracy}
\label{fig:5shot_16task_tiered_meta}
\end{minipage}

\caption{Results on the 1- and 5-shot, 5-way 16-task setup with a 4-Conv backbone and ProtoNets meta-learner. Evaluations are on CIFAR100, Mini-ImageNet and Tiered-ImageNet datasets. 
}
\label{fig:16_task_3dataset}
\end{figure*}

Here we describe the experimental results on long/short task sequences and comparison with continual learning methods.
\minisection{Experiments on long task sequences.}
In Fig.~\ref{fig:16_task_3dataset} we report results on 16-task 1-shot/5-shot 5-way incremental meta-learning for three
datasets: CIFAR100, Mini-ImageNet and Tiered-ImageNet (Results on CUB dataset in supplementary). The first and third columns in
Fig.~\ref{fig:16_task_3dataset} show mean accuracy on previous tasks, from
which it is obvious that we achieve significantly less forgetting compared to other
methods. From the second and fourth columns, the meta-test accuracy for ERD
increases with more meta-training tasks due to seeing more diverse classes,
while for IDA and EIML, in some training sessions, the performance drops
significantly. This might be due to forgetting on previous tasks and overfitting
 on current task. Notably, for our method the meta-test accuracy after the
last task is much closer to the joint training upper bound, which is learned
using all training tasks.

Note also that EIML works much better than IDA after about the first 4 tasks.
That might be for the same reason that, in the original IDA paper, they report
similar results for both IDA and EIML, which is simply due to evaluating on only
very short sequences of two or three tasks. This is likely due to anchor drift
in IDA and the fact that in EIML exemplars could be used to re-calibrate them.
In general, all methods work better in the 5-shot. The underlying reason for
this is that the 5-shot is simpler compared to 1-shot.

\minisection{Experiments on short task sequences.}
\begin{table}[t]
\begin{center}
\scalebox{0.58}{
\begin{tabular}{|r|cccc|cccc|cccc|}
\hline

Learner: & \multicolumn{12}{|c|}{ProtoNets}  \\
\hline

Dataset:& \multicolumn{4}{|c|}{Mini-ImageNet}   & \multicolumn{4}{|c|}{CIFAR100 }  & \multicolumn{4}{|c|}{ CUB} \\
\hline

Backbone: & \multicolumn{12}{|c|}{4-Conv }   \\

\hline

& \multicolumn{4}{|c|}{Upper bound: 53.2} & \multicolumn{4}{|c|}{Upper bound: 55.4} & \multicolumn{4}{|c|}{Upper bound: 61.1} \\
\hline
 Sessions: & 1&2&3&4    & 1&2&3&4   & 1&2&3&4 \\
\hline\hline

FT &43.8& 	44.1 & 	42.4 & 	37.9 &   44.6 & 45.1 & 48.0 & 45.5  &    45.1 & 54.6 & 54.9 & 58.8     \\  
IDA &43.8&  48.3  & 47.2 & 42.3    &     44.6 & 48.0 & 51.3 & 47.6 &   45.1 & \textbf{54.7} & 54.9 & 58.7   \\  
EIML  &43.8 &  48.8  &  49.4  & 47.5 & 44.6 & 48.0 & 52.0 & 51.7 & 45.1 & 53.4 & 55.0& 58.9   \\

ERD  &43.8   &\textbf{51.1}  &\textbf{52.3}  &\textbf{53.0}  &  44.6 & \textbf{49.5} & \textbf{53.6} & \textbf{55.1} &  45.1 & 53.9 & \textbf{58.3} & \textbf{60.8}    \\
\hline\hline

Backbone: & \multicolumn{12}{|c|}{ResNet-12} \\
\hline
& \multicolumn{4}{|c|}{Upper bound: 59.9} & \multicolumn{4}{|c|}{Upper bound: 61.8} & \multicolumn{4}{|c|}{Upper bound: 74.8 } \\
\hline
 Sessions: & 1&2&3&4  & 1&2&3&4&  1&2&3&4 \\
 \hline\hline
 FT&       45.7 & 45.9 & 42.1 & 37.7   & 47.0 & 45.0 & 51.0 & 44.6 & 53.4 & 64.0 & 63.7 & 66.8\\
 IDA&  45.7 & 53.0 & 53.7 & 47.6 & 47.0 & 53.6 & 59.2 & 54.8 & 53.4  & 64.4 & 68.8 & 73.3 \\
 EIML&  45.7 & 53.2 & 56.5 & 55.8 & 47.0 & 53.3 & 58.3 & 57.8 & 53.4  & 62.8 & 69.1 & 73.3\\
 ERD&    45.7 & \textbf{55.2} & \textbf{58.2} & \textbf{59.3} &  47.0 & \textbf{55.6} & \textbf{61.3} & \textbf{61.4}  & 53.4  & \textbf{66.1} & \textbf{72.4} & \textbf{74.1} \\
 \hline
 
\end{tabular}
}
\end{center}
\caption{Meta-test accuracy by training session in the 4-task setting. We evaluate \textit{1-shot} and \textit{5-way} few-shot recognition on three datasets using two different backbones. 
}
\label{tab:complete_table_3datasets}
\end{table}
We also compare our method with others on short sequences.
In table~\ref{tab:complete_table_3datasets}, we first evaluate our model with a
4-Conv backbone on 1-shot/5-way few-shot on three datasets (5-shot results in supplementary). We see that FT suffers from
catastrophic forgetting, and meta-test accuracy drops dramatically and exhibits
overfitting to the current task. IDA is not able to improve meta-test accuracy
on Mini-ImageNet, but improves performance on CIFAR100 and CUB. As for EIML,
with exemplars it shows large improvement compared to IDA. However, our method
ERD outperforms EIML by a large margin after learning all four tasks. These
results further confirm the observations on the 16-task setting. ERD not
only achieves the best performance with less forgetting, but also gets
closer to the upper bound after the last task. Note also that CIFAR100
and Mini-ImageNet are coarse-grained datasets, compared to CUB, which makes few
shot classification much harder due to intra-class variability.

Finally, we consider ResNet-12 as a backbone to show that ERD can be
applied to different backbones. Our method achieves
consistently better performance over other methods with much higher accuracy than using 4-Conv backbone.

\minisection{Comparison with standard CL methods.} 
\label{sec:benchmark}
As shown in Table~\ref{tab:compare_CL_methods}, we compare our method with three state-of-the-art CL methods: iCaRL~\cite{rebuffi2017icarl}, PODNet~\cite{douillard2020podnet} and UCIR~\cite{hou2019learning}. For the evaluation on seen classes, we followed the same protocol as IDA, where the average classification accuracy is calculated over $N_{ep}$ episodes. Note that this is different from the evaluation in regular CL. Thus, these methods cannot be directly applied in this scenario. Instead, we use them to continually learn representations and then evaluate them with a nearest centroid classifier for few-shot learning. Observe how, on seen classes evaluation, UCIR works better than iCaRL and PODNet. But under meta-test evaluation, iCaRL works the best among them. PODNet performs similar to the FT baseline in both cases. Clearly, without episodic training, regular CL methods are inferior to our proposed ERD.

\begin{table}[h]
\begin{center}
\scalebox{0.68}{
\begin{tabular}{|r|ccccc|ccccc|}

\hline
\multicolumn{11}{|c|}{Dataset: CIFAR100 , Learner: ProtoNets, Backbone: 4-Conv} \\
\hline
Evaluation:  & \multicolumn{5}{|c|}{Meta-test accuracy} & \multicolumn{5}{|c|}{Mean accuracy on seen classes} \\ 
\hline
 Sessions:     & 2&4&8&16 & avg  & 2&4&8&16 & avg  \\
\hline\hline
FT     & 37.3 & 37.6 & 40.0 & 34.4 & 38.1 & 44.8 & 44.1 & 40.9 & 37.8 & 42.5  \\ 
IDA    & 39.8 & 39.3 & 42.2 & 35.9 & 40.3 & 50.6 & 46.2 & 44.1 & 39.6 & 44.7  \\ 
EIML   & 39.7 & 40.2 & 44.9 & 42.0 & 43.1 & 51.1 & 48.9 & 46.8 & 46.7 & 48.6  \\ 
ERD    & \textbf{40.1} & \textbf{45.0} & \textbf{50.0} & \textbf{50.5} & \textbf{48.1} & {51.0} & \textbf{52.2} & \textbf{53.7} & \textbf{54.8} & \textbf{53.9}  \\ 
\hline
iCaRL  & 39.0 & 42.0 & 43.4 & 45.2 & 43.1 & 50.1 & 47.2 & 46.5 & 48.0 & 48.5  \\ 
UCIR   & 35.1 & 36.3 & 39.5 & 42.2 & 39.3 & \textbf{53.6} & 50.5 & 50.1 & 51.9 & 52.4  \\ 
PODNet & 36.0 & 37.0 & 37.1 & 36.4 & 37.0 & 52.9 & 43.8 & 41.0 & 41.1 & 44.6  \\

\hline
\end{tabular}
}
\end{center}
\caption{Meta-test accuracy and mean accuracy on seen classes by training session on the 16-task setting. We evaluate 1-shot/5-way few-shot recognition to compare with continual learning methods (5-shot results in supplementary).}
\label{tab:compare_CL_methods}
\end{table}

\subsection{Ablation study} \label{subsec:ablation_study} 
\begin{figure*}[h]
\begin{minipage}[b]{0.24\linewidth}
\centering
\includegraphics[width=\textwidth]{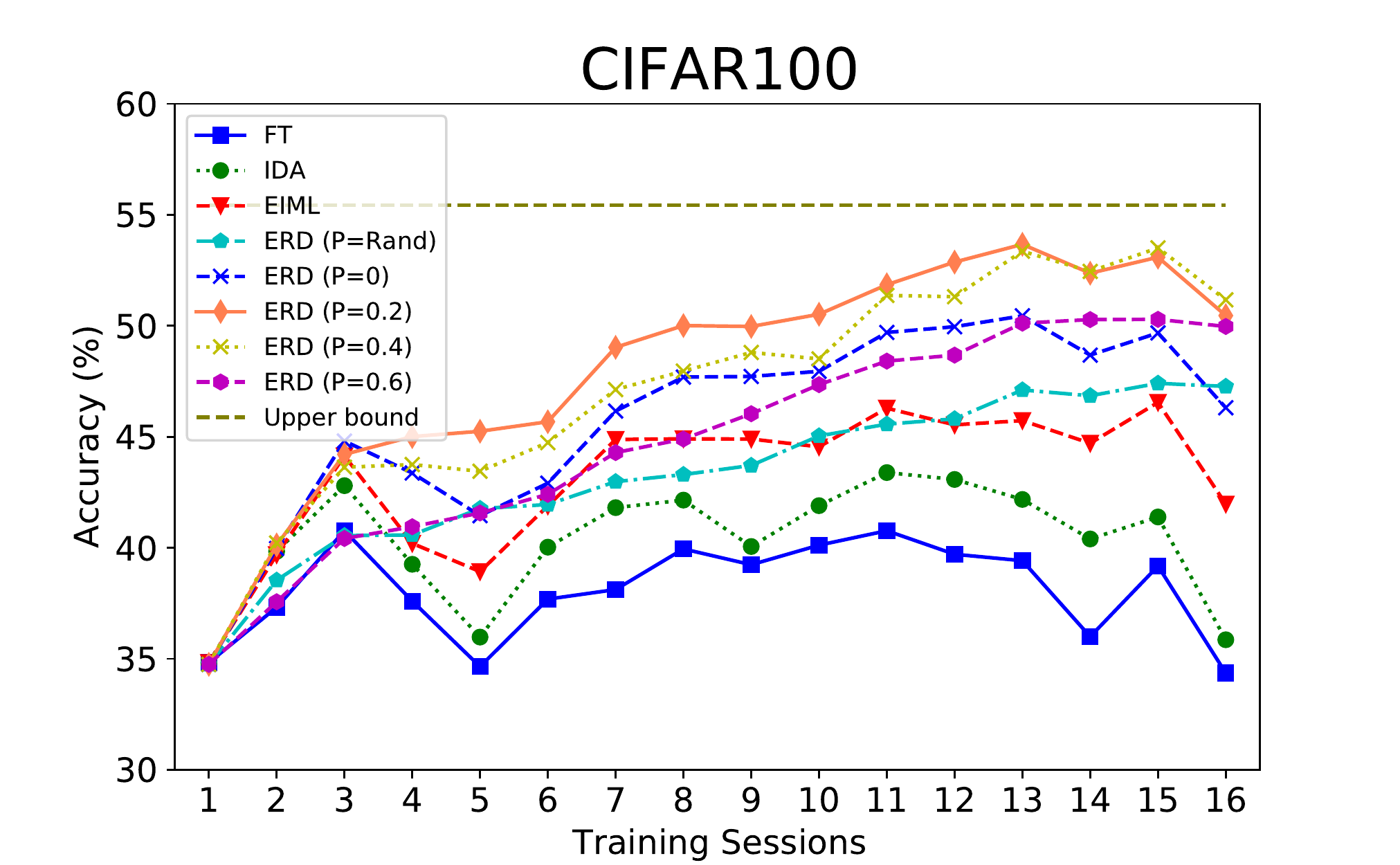}
\subcaption{Ablation on $P$}
\label{fig:1shot_16task_cifar_ablate_P}
\end{minipage}
\begin{minipage}[b]{0.24\linewidth}
\centering
\includegraphics[width=\textwidth]{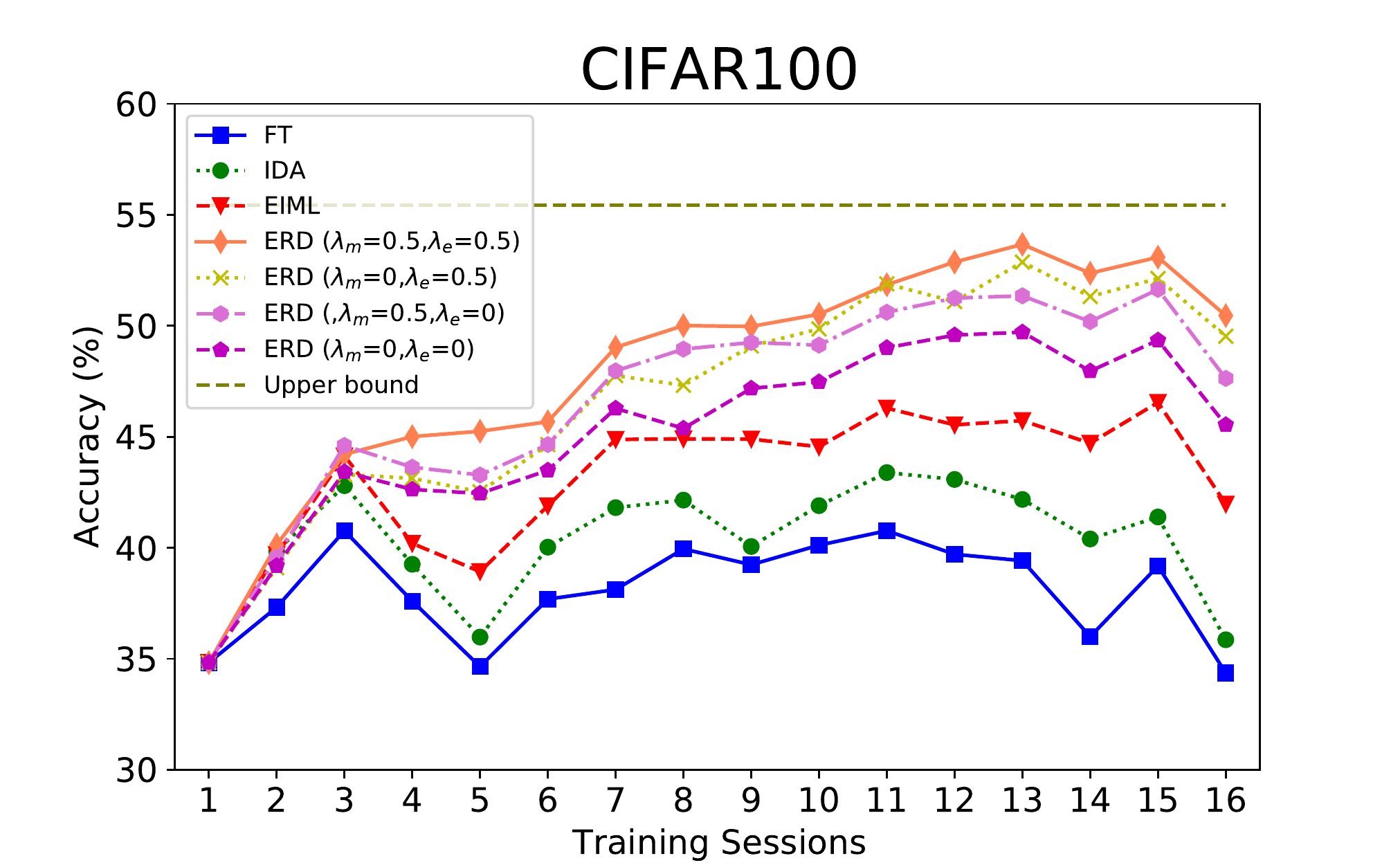}
\subcaption{Ablation on $\lambda_e$, $\lambda_m$}
\label{fig:1shot_16task_cifar_ablate_lam2}
\end{minipage}
\begin{minipage}[b]{0.24\linewidth}
\centering
\includegraphics[width=\textwidth]{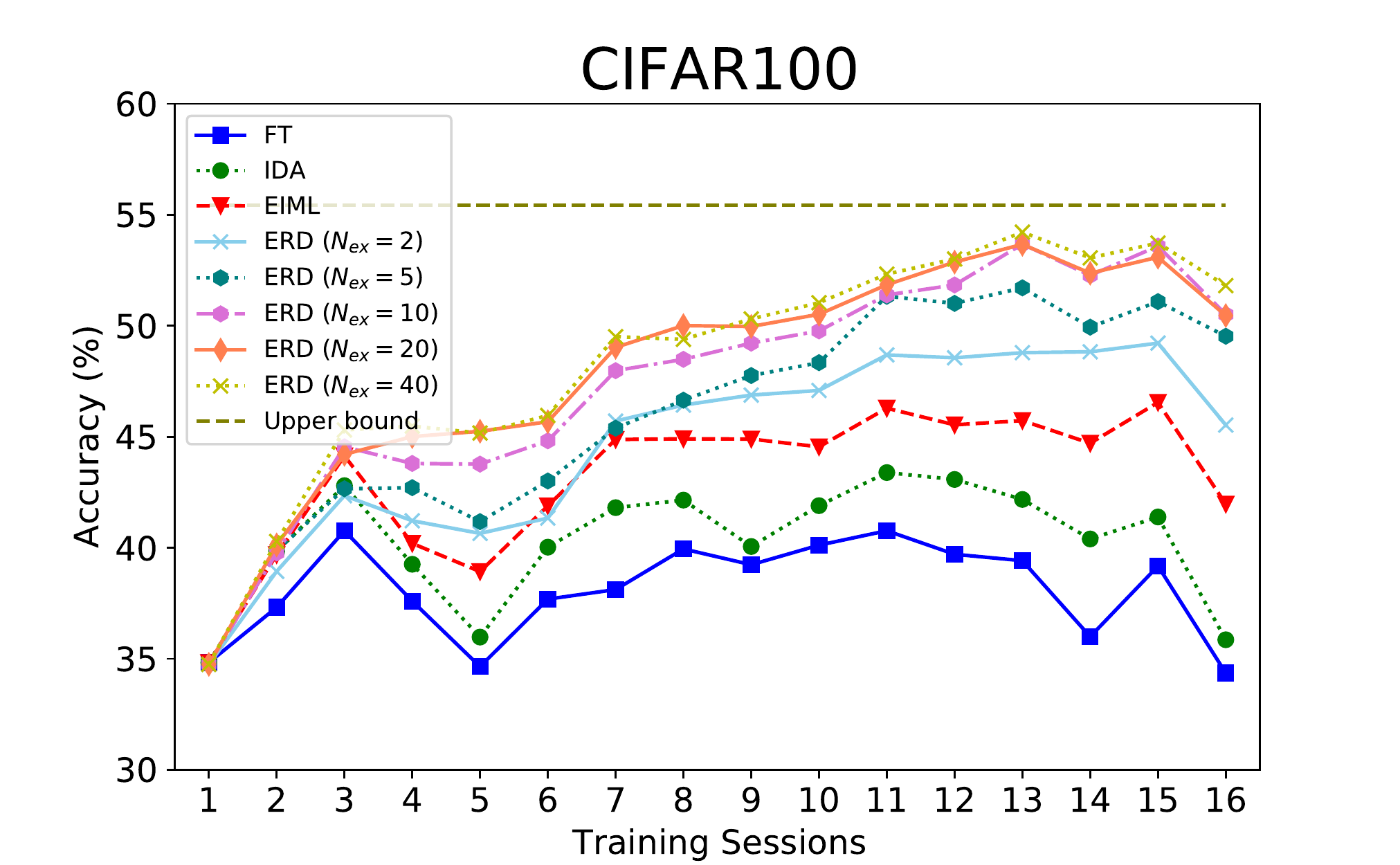}
\subcaption{Ablation on $N_{ex}$}
\label{fig:1shot_16task_cifar_ablate_exem_inc}
\end{minipage}
\begin{minipage}[b]{0.24\linewidth}
\centering
\includegraphics[width=\textwidth]{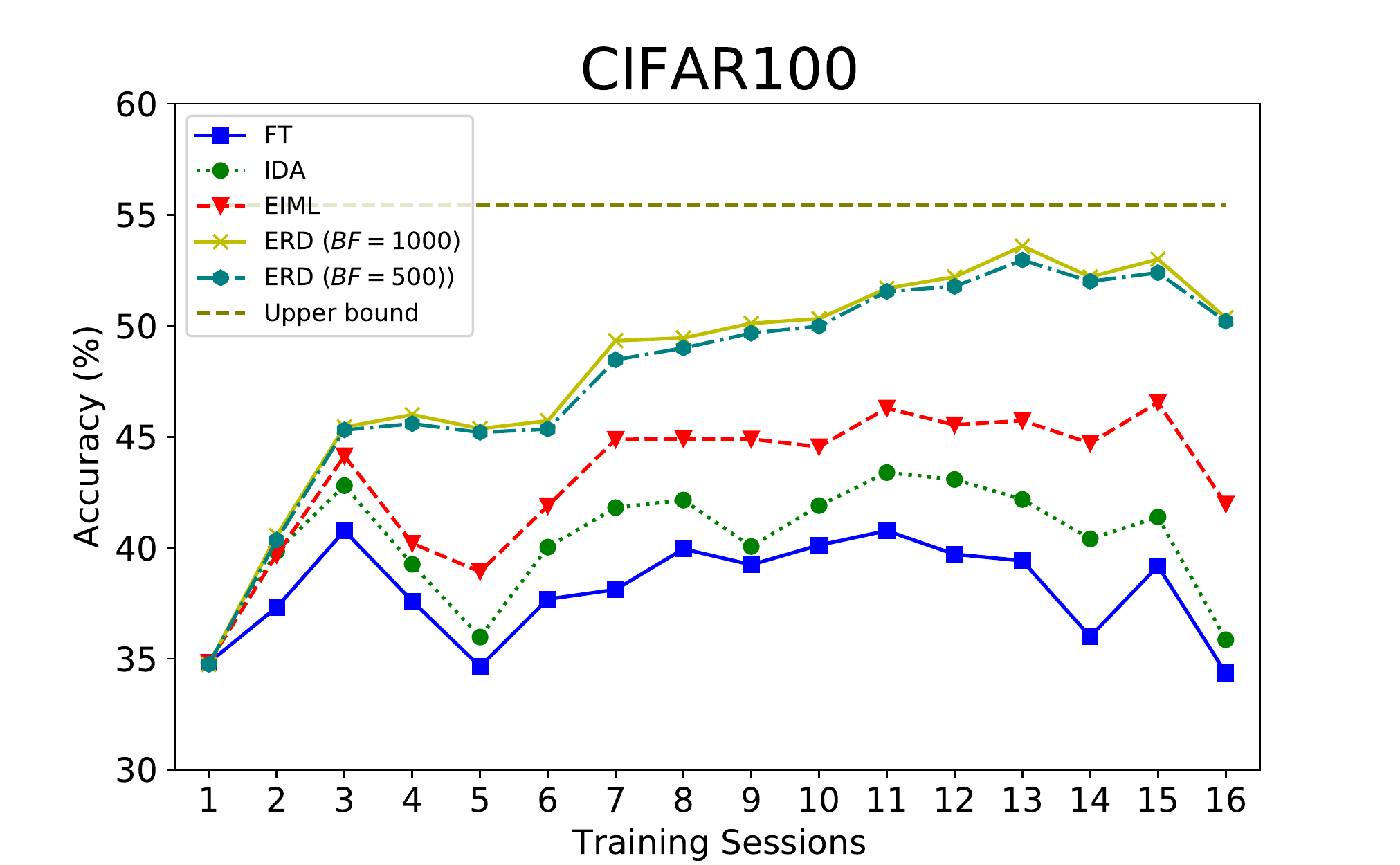}
\subcaption{Ablation on $BF$}
\label{fig:1shot_16task_cifar_ablate_exem_bound}
\end{minipage}
\caption{Ablation study on 16-task 1-shot/5-way setup on CIFAR100 with 4-Conv. We plot the meta-test accuracy to compare.
}
\label{fig:16_task_cifar_ablation}
\end{figure*}

In
Fig.~\ref{fig:16_task_cifar_ablation} we show the ablation study
on CIFAR100 under the 16-task 1-shot/5-way setup with 4-Conv as the backbone. We plot \textit{meta-test accuracy} curves to compare among variants since it is the most important evaluation metric in IML.

\minisection{Ablation on $P$ with $\lambda_m=\lambda_e=0.5$.} As shown in
Fig.~\ref{fig:1shot_16task_cifar_ablate_P}, ERD obtains the best performance
with $P=0.2$. This is what we use by default for all previous experiments. When
$P=0$, it means there are no previous classes in the \textit{cross-task sub-episode}, which
performs worse than our variants with higher probabilities, especially with
$P=0.2$ and $P=0.4$. As $P$ decreases from 0.6 to 0.2, the performance
consistently improves. The reason is that lower $P$ means having more current
samples, which can ensure the diversity of training samples. This phenomenon is
different from the conventional use of exemplars in incremental learning, where
more balanced exemplar sampling is preferable. $P=\mbox{Rand}$ instead means $P$
is not fixed, but that classes in each cross-task sub-episode are randomly selected from all
encountered classes up to now and $P$ is increasing with successive tasks. It
achieves worse results because there are more and more previous
classes with less diverse samples. Most of our variants outperform EIML by a
large margin. We keep $P \geq 0.2$ to ensure that at least one previous class occurs
in each episode for 5-way few-shot learning.

As a conclusion, the value of $P$ controls the number of previous classes used in the current episode, which is a key component of cross-task episodic training.

\minisection{Ablation on $\lambda_{m}$ and $\lambda_{e}$ with $P=0.2$.} To
understand the role of each distillation component in
Eq~(\ref{eq:erd_loss}), we ablate the distillation loss terms. As shown in
Fig.~\ref{fig:1shot_16task_cifar_ablate_lam2}, our method achieves the best with $\lambda_{m}=0.5$ and $\lambda_{e}=0.5$, which indicates both
distillation terms playing a crucial role in overcoming forgetting and
generalizing to unseen tasks. ERD with $\lambda_{m}=0.5$, $\lambda_{e}=0$
works similarly to ERD with $\lambda_{m}=0$, $\lambda_{e}=0.5$. They both
achieve much better performance than without using distillations
($\lambda_{m}=0$ and $\lambda_{e}=0$).

\minisection{Ablation on memory buffer with $P=0.2, \lambda_m=\lambda_e=0.5$.} In
this experiment, we fix other hyper-parameters to show how different numbers of
exemplars affect incremental learning performance.
We give ablations on $N_{ex}$ and with bounded buffer size $BF$ which are both commonly used for exemplar rehearsal. 
From
Fig.~\ref{fig:1shot_16task_cifar_ablate_exem_inc} we see that increasing
$N_{ex}$ leads to a noticeable increase in performance going from
2 to 20 exemplars. However, the gain is marginal beyond
20 exemplars per class.
 From Fig.~\ref{fig:1shot_16task_cifar_ablate_exem_bound}, we can observe that with a smaller bounded buffer with only $BF=500$ exemplars, ERD is still close to the joint training upper bound, showing the importance of proposed sub-episodes.

\subsection{Extension to Relation Networks} \label{subsec:extension_relation_nets_expr}

Since in Relation Networks there is no embedding to
exploit for computing prototypes as in ProtoNets,
IDA and EIML cannot be directly applied. Therefore we only compare with FT in
this experiment. As the experimental results shown in
Table~\ref{tab:complete_table_3datasets_relationnets}, our model not only surpasses the FT baseline significantly, but
also gets close to the joint training upper bounds after the last task,
especially on the CUB dataset.

\begin{table}[htbp!]
\begin{center}
\scalebox{0.55}{
\begin{tabular}{|r|cccc|cccc|cccc|}
\hline

Learner: & \multicolumn{12}{|c|}{Relation Networks} \\
\hline
Datasets:  & \multicolumn{4}{|c|}{ Mini-ImageNet} & \multicolumn{4}{|c|}{CIFAR100}  &\multicolumn{4}{|c|}{CUB} \\
\hline

Backbone:   & \multicolumn{4}{|c|}{ 4-Conv} & \multicolumn{4}{|c|}{4-Conv } & \multicolumn{4}{|c|}{4-Conv}\\
\hline

\multicolumn{13}{|c|}{1-shot/5-way \textbf{\textit{16-task}} setting}  \\
\hline

 & \multicolumn{4}{|c|}{Upper bound: 52.0} & \multicolumn{4}{|c|}{Upper bound: 59.2 } & \multicolumn{4}{|c|}{Upper bound: 51.6}\\
\hline
 Sessions:    & 2&4&8&16  & 2&4&8&16  & 2&4&8&16   \\
\hline\hline
FT &   24.4 & 28.5 & 25.5 & 28.7 & 31.1 & 30.0 & 35.2 & 26.8  & 37.1 & 38.1 & 37.0 & 34.0\\

ERD &  \textbf{27.7} & \textbf{29.9} & \textbf{34.5} & \textbf{30.1}  &  \textbf{35.6} & \textbf{39.3}  & \textbf{45.7}  & \textbf{35.9} & \textbf{37.3} & \textbf{42.9} & \textbf{47.9} & \textbf{42.5} \\

\hline\hline
 \multicolumn{13}{|c|}{1-shot/5-way \textbf{\textit{4-task}} setting} \\
\hline

 & \multicolumn{4}{|c|}{Upper bound: 52.0} & \multicolumn{4}{|c|}{Upper bound: 59.2 } & \multicolumn{4}{|c|}{Upper bound: 51.6}\\
\hline
 Sessions: & 1&2&3&4    & 1&2&3&4  &1&2&3&4  \\
\hline\hline
FT  & 41.70 & 41.65 & 38.51  & 33.33 & 42.9 & 45.3 & 45.7 & 42.3 & 45.7 & 47.9 & 48.5 & 50.3 \\
ERD & 41.70 & \textbf{45.84} & \textbf{48.35} & \textbf{49.73} & 42.9 &  \textbf{48.2} &  \textbf{51.2} &  \textbf{51.4} & 45.7 &  \textbf{48.5} &  \textbf{49.4} &  \textbf{51.4}  \\

\hline

\end{tabular}
}
\end{center}
\caption{Meta-test accuracy by training sessions on the 4-task and 16-task settings. We evaluate 1-shot/5-way few-shot recognition on Mini-ImageNet, CIFAR-100 and CUB.}
\label{tab:complete_table_3datasets_relationnets}
\end{table}

\section{Conclusions} \label{sec:conclusion}

In this paper we proposed Episodic Replay Distillation, an approach to
incremental few-shot recognition that uses episodic meta-learning over episodes
split into cross-task and exemplar sub-episodes. This division into two types of
few-shot learning sub-episodes allows us to balance acquisition of new task
knowledge with retention of knowledge from previous tasks and thus avoid
catastrophic forgetting. Experiments on multiple few-shot learning datasets
demonstrate the effectiveness of ERD. Our method is especially effective on long
task sequences, where we significantly close the gap between incremental
few-shot learning and the joint training upper bound.

\minisection{Acknowledgments.} We acknowledge the support from Huawei Kirin Solution.

\bibliography{longstrings,mybib}

\newpage\hbox{}\thispagestyle{empty}\newpage

\includepdf[pages={{},-}]{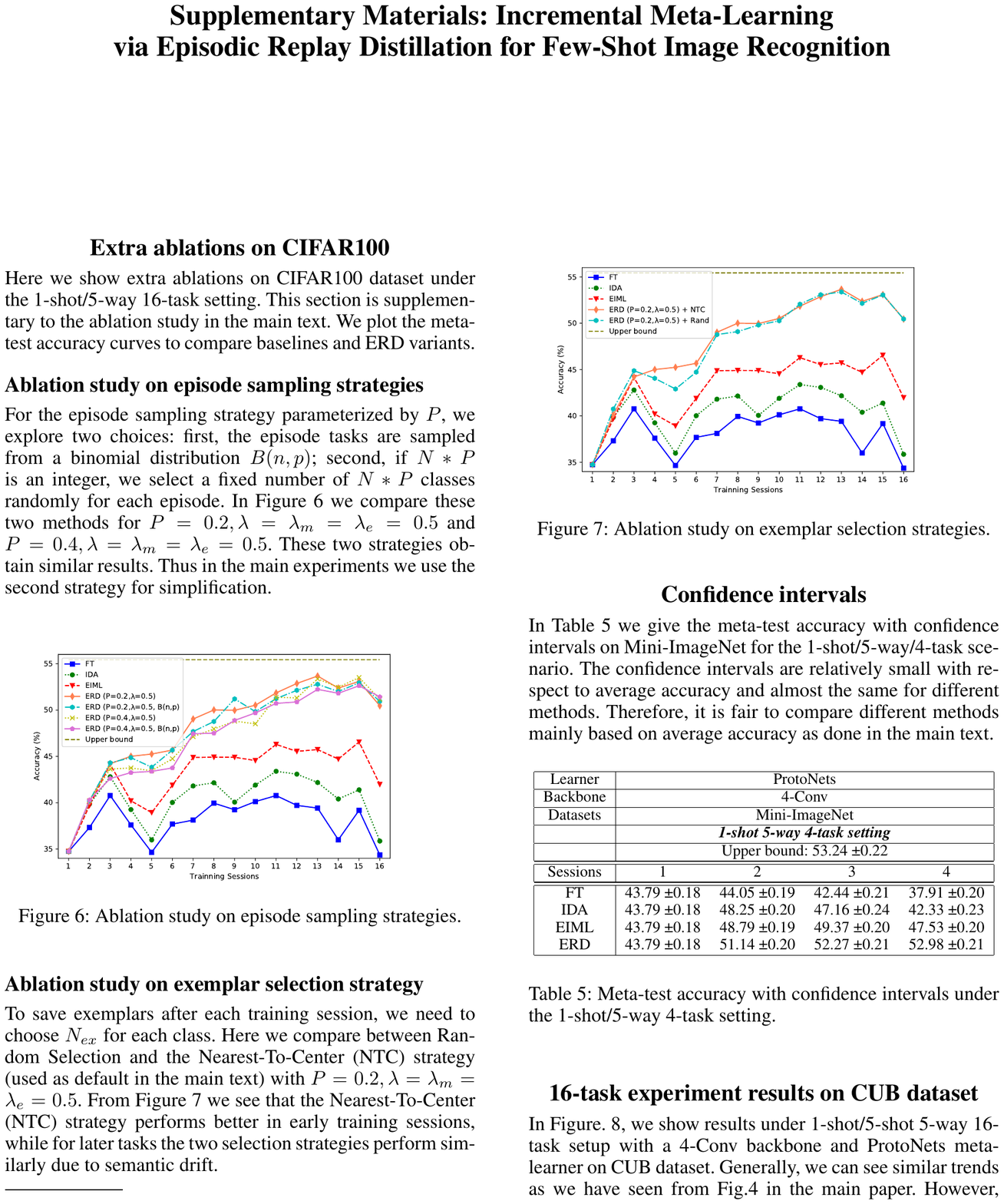}

\end{document}


\maketitle

\setcounter{figure}{5}
\setcounter{table}{4}
\setcounter{equation}{10}

\section{Extra ablations on CIFAR100}

Here we show extra ablations on CIFAR100 dataset under the 1-shot/5-way 16-task setting. This section is supplementary to the ablation study in the main text. We plot the meta-test accuracy curves to compare baselines and ERD variants.

\subsection{Ablation study on episode sampling strategies}

\begin{figure}[h]
\begin{minipage}[b]{0.91\linewidth}
\centering
\includegraphics[width=\textwidth]{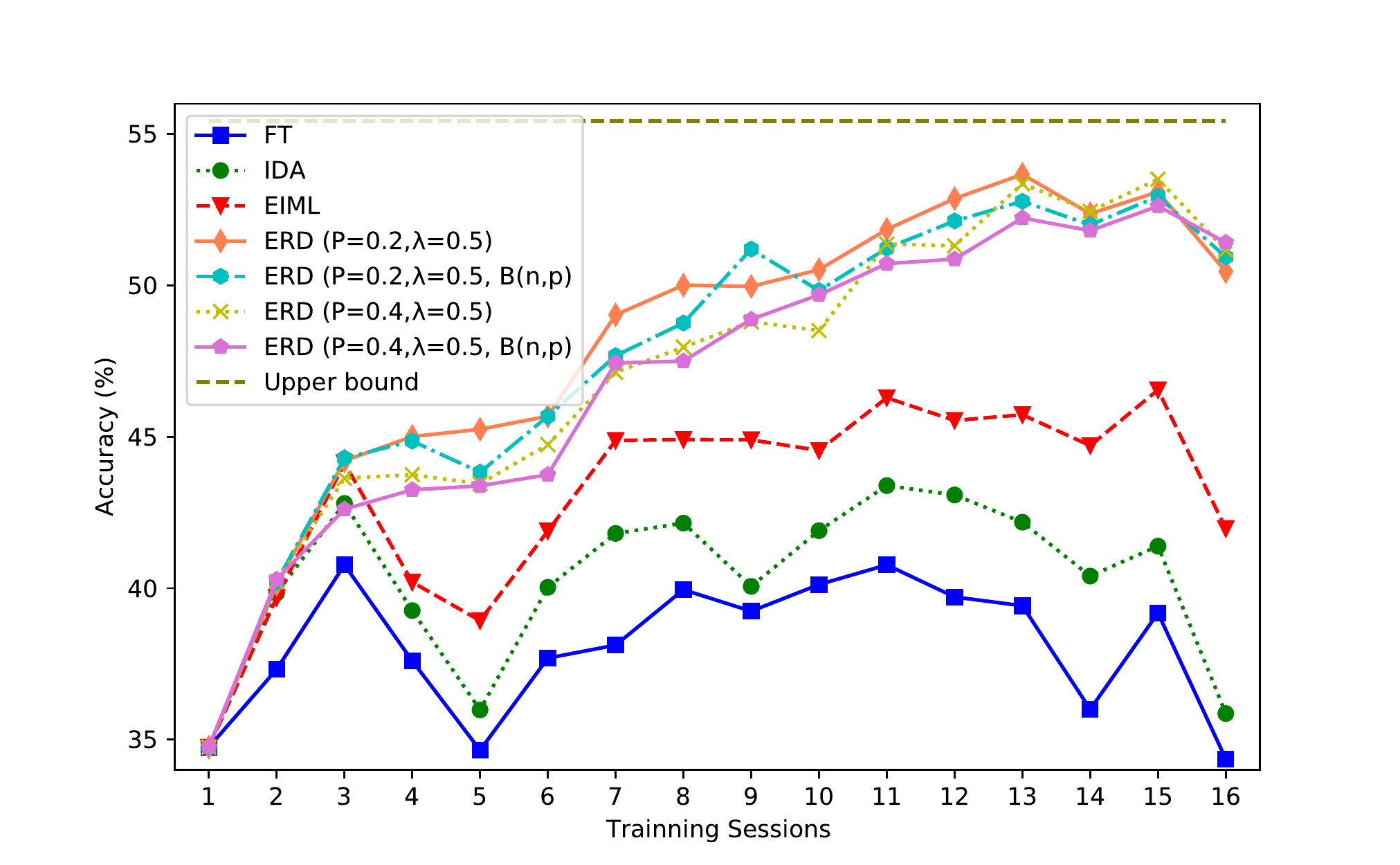}
\end{minipage}
\caption{Ablation study on episode sampling strategies.
}
\label{fig:1shot_16task_cifar_ablate_strategy}
\end{figure}

For the episode sampling strategy parameterized by $P$, we explore two choices: first, the episode tasks are sampled from a binomial distribution $B(n,p)$; second, if $N*P$ is an integer, we select a fixed number of $N*P$ classes randomly for each episode. In Figure~\ref{fig:1shot_16task_cifar_ablate_strategy} we compare these two methods for $P=0.2, \lambda=\lambda_m=\lambda_e=0.5$ and $P=0.4, \lambda=\lambda_m=\lambda_e=0.5$. These two strategies obtain similar results. Thus in the main experiments we use the second strategy for simplification.

\subsection{Ablation study on exemplar selection strategy}

\begin{figure}[h]
\begin{minipage}[b]{0.91\linewidth}
\centering
\includegraphics[width=\textwidth]{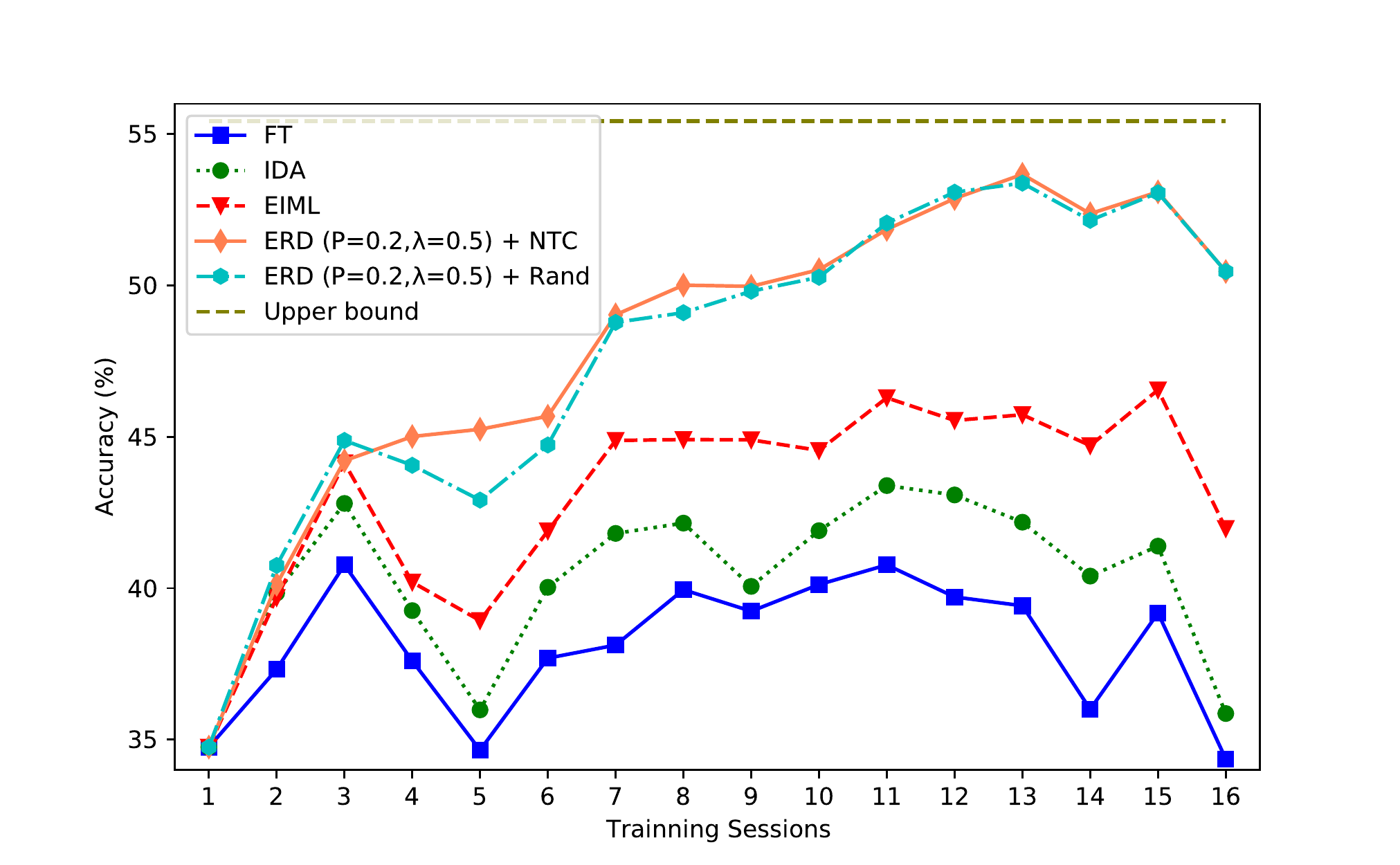}
\end{minipage}
\caption{Ablation study on exemplar selection strategies.
}
\label{fig:1shot_16task_cifar_ablate_exemplar_selection}
\end{figure}

To save exemplars after each training session, we need to choose $N_{ex}$ for each class. Here we compare between Random Selection and the Nearest-To-Center (NTC) strategy (used as default in the main text) with $P=0.2, \lambda=\lambda_m=\lambda_e=0.5$. From Figure~\ref{fig:1shot_16task_cifar_ablate_exemplar_selection} we see that the Nearest-To-Center (NTC) strategy performs better in early training sessions, while for later tasks the two selection strategies perform similarly due to semantic drift.

\section{Confidence intervals}
In Table~\ref{tab:interva_confidence} we give the meta-test accuracy with confidence intervals on Mini-ImageNet for the 1-shot/5-way/4-task scenario. The confidence intervals are relatively small with respect to average accuracy and almost the same for different methods. Therefore, it is fair to compare different methods mainly based on average accuracy as done in the main text.

\begin{table}[H]
\begin{center}
\scalebox{0.8}{
\begin{tabular}{|c|cccc|cccc|cccc|cccc|}
\hline

Learner & \multicolumn{4}{|c|}{ProtoNets}  \\
\hline
Backbone  & \multicolumn{4}{|c|}{4-Conv}  \\
\hline
Datasets & \multicolumn{4}{|c|}{Mini-ImageNet}   \\
\hline

& \multicolumn{4}{|c|}{\textbf{\textit{1-shot 5-way 4-task setting}}}\\
\hline

& \multicolumn{4}{|c|}{Upper bound: 53.24 \textpm 0.22}   \\
\hline\hline
 Sessions & 1&2&3&4    \\
\hline\hline
FT & 43.79 \textpm 0.18 & 44.05  \textpm 0.19 & 42.44 \textpm 0.21 & 37.91 \textpm 0.20 \\  
IDA & 43.79 \textpm 0.18 & 48.25 \textpm 0.20 & 47.16 \textpm 0.24 & 42.33 \textpm 0.23 \\  
EIML &  43.79 \textpm 0.18 & 48.79 \textpm 0.19 & 49.37 \textpm 0.20 & 47.53 \textpm 0.20 \\
ERD  & 43.79 \textpm 0.18 & 51.14 \textpm 0.20 & 52.27 \textpm 0.21 &  52.98 \textpm 0.21 \\
\hline

\end{tabular}
}
\end{center}
\caption{Meta-test accuracy with confidence intervals under the 1-shot/5-way 4-task setting.}
\label{tab:interva_confidence}
\end{table}

\begin{figure*}[h]
\begin{minipage}{0.48\linewidth}
\centering
\includegraphics[width=\textwidth]{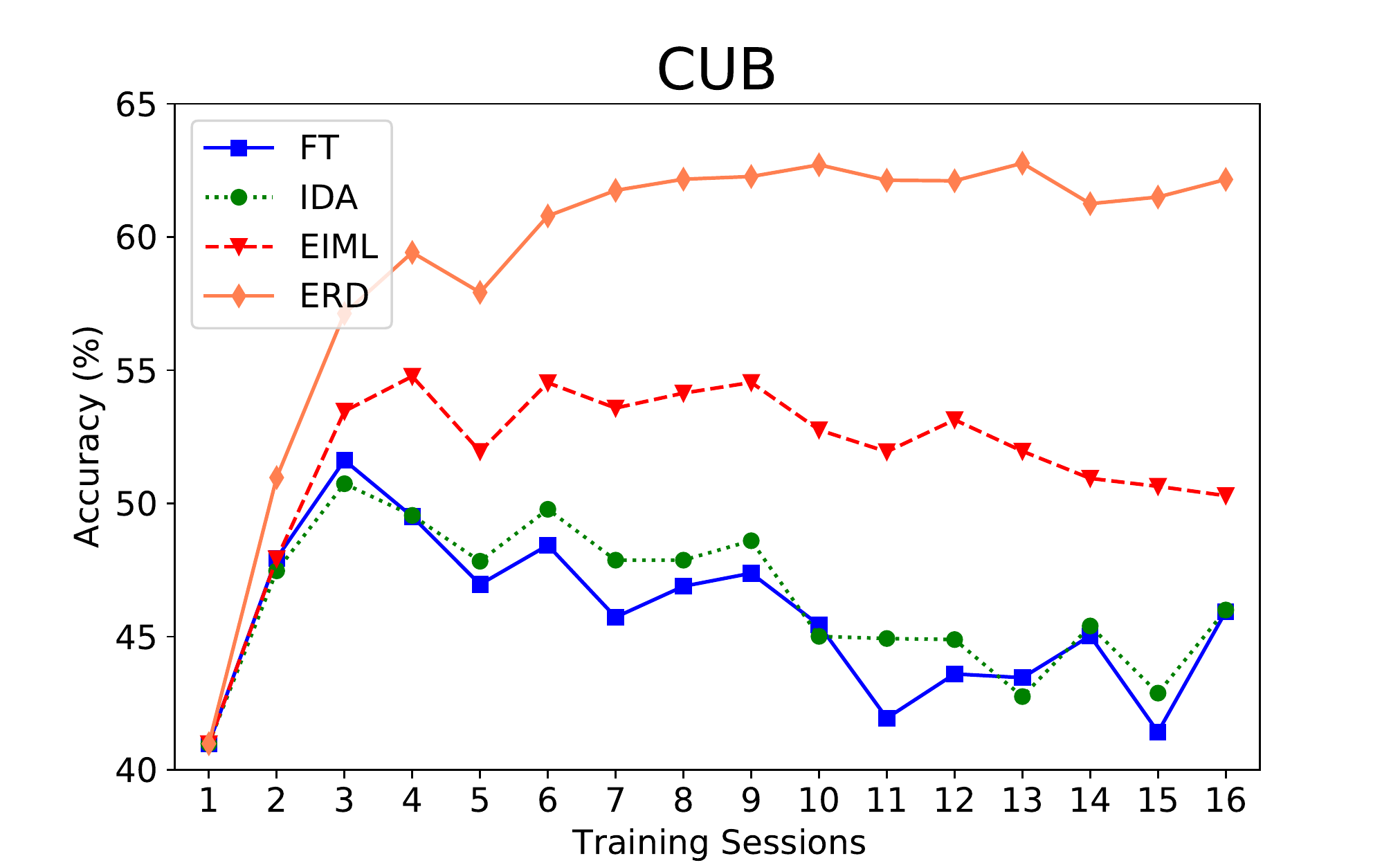}
\subcaption{1-shot mean accuracy on seen classes}
\label{fig:1shot_16task_cub_seen}
\end{minipage}
\begin{minipage}{0.48\linewidth}
\centering
\includegraphics[width=\textwidth]{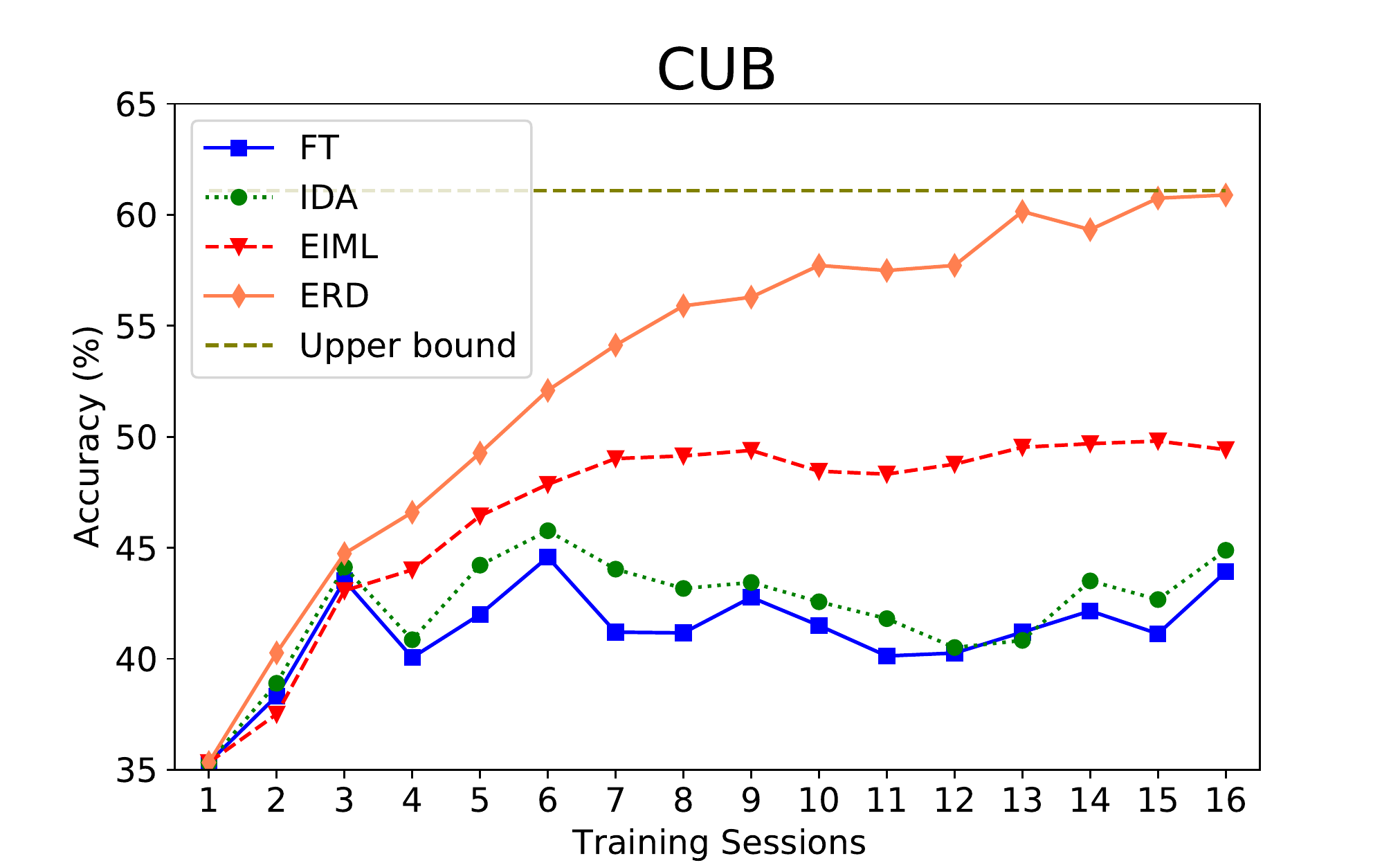}
\subcaption{1-shot meta-test accuracy}
\label{fig:1shot_16task_cub_meta}
\end{minipage}
\begin{minipage}{0.48\linewidth}
\centering
\includegraphics[width=\textwidth]{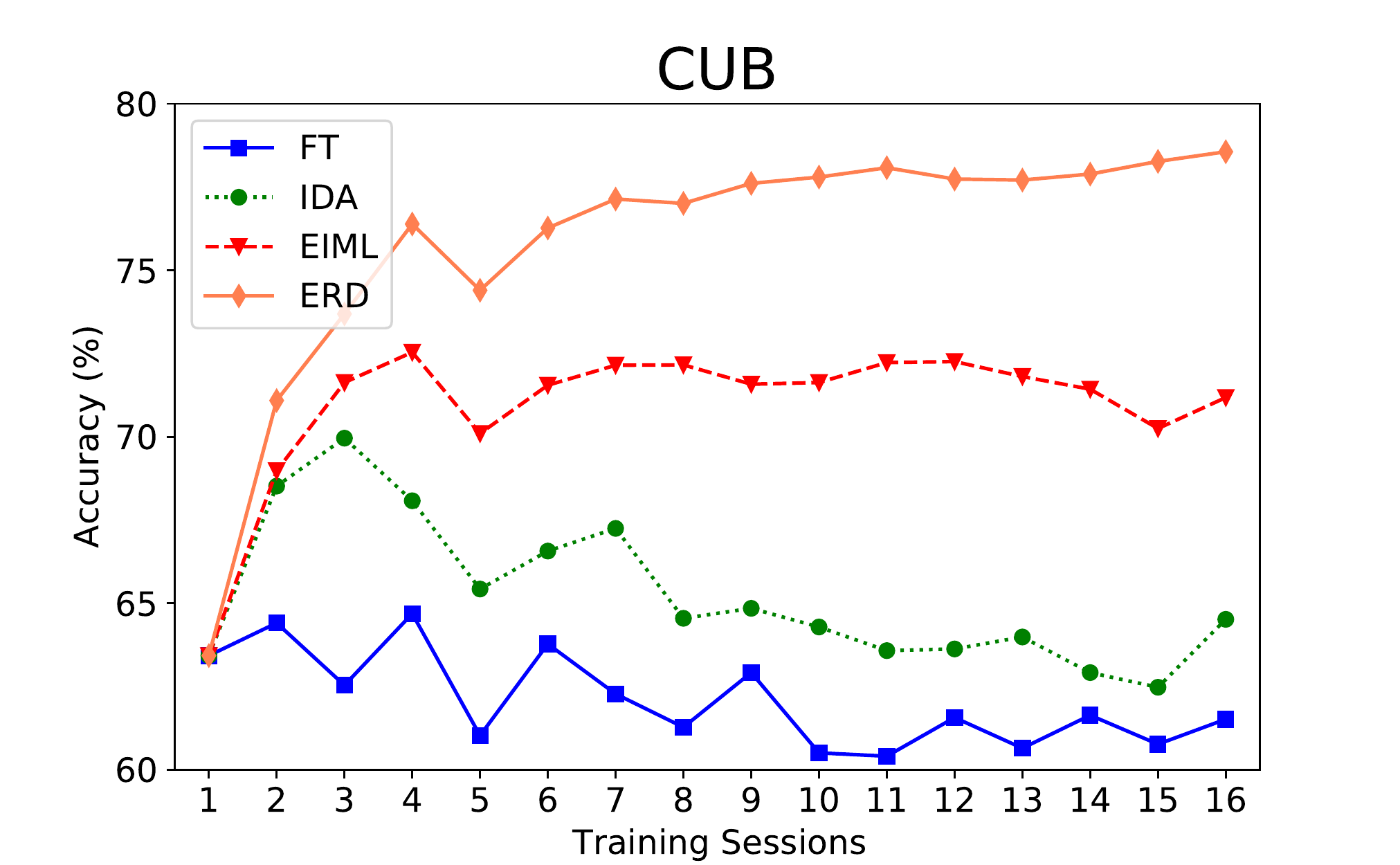}
\subcaption{5-shot mean accuracy on seen classes}
\label{fig:5shot_16task_cub_seen}
\end{minipage}
\hfill
\begin{minipage}{0.48\linewidth}
\centering
\includegraphics[width=\textwidth]{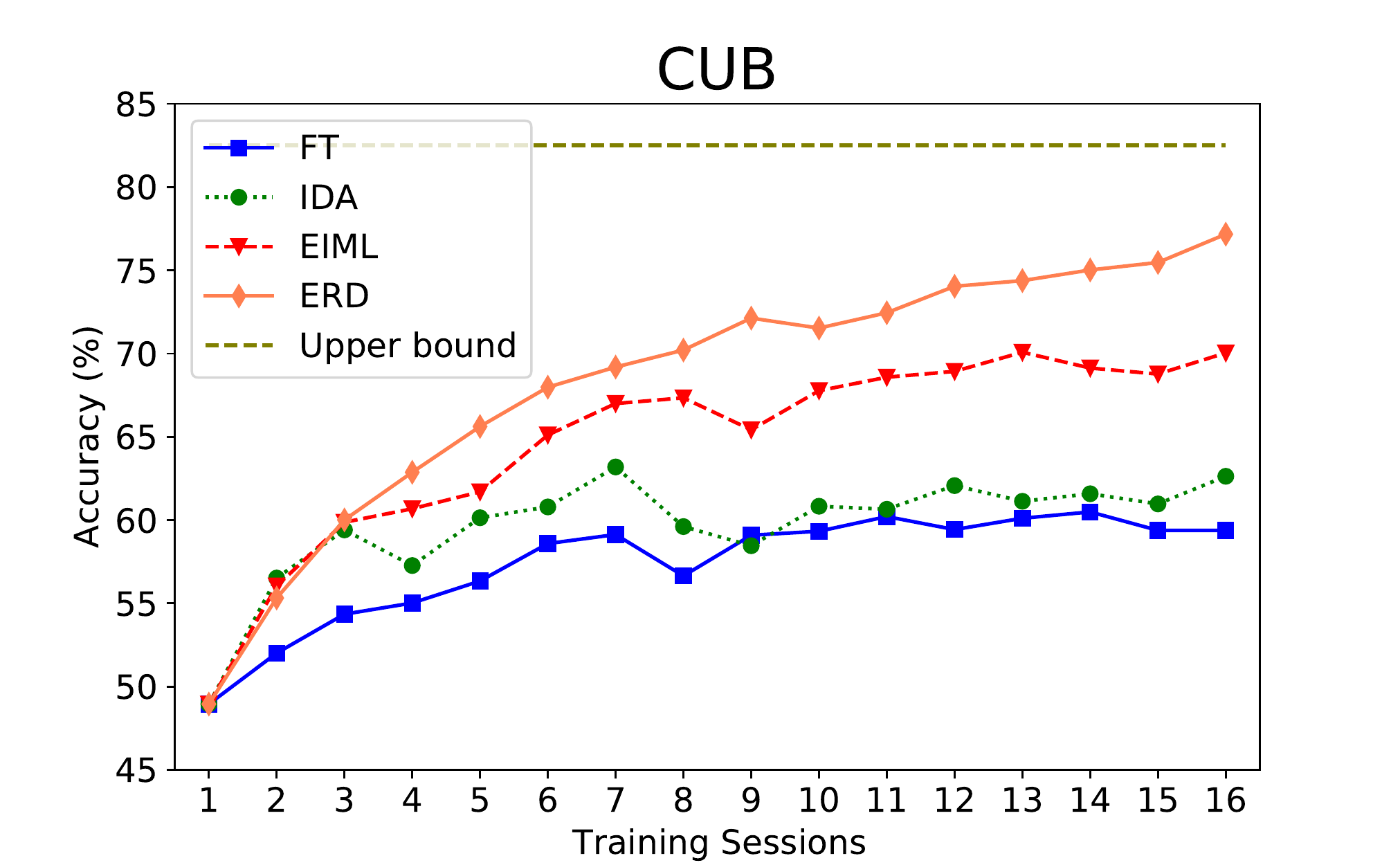}
\subcaption{5-shot meta-test accuracy}
\label{fig:5shot_16task_cub_meta}
\end{minipage}
\caption{Results on CUB dataset under 1-shot/5-shot 16-task setting with a 4-Conv backbone and ProtoNets meta-learner.}
\label{fig:16_task_cub}
\end{figure*}

\section{16-task experiment results on CUB dataset}
In Figure.~\ref{fig:16_task_cub}, we show results under 1-shot/5-shot 5-way 16-task setup with a 4-Conv backbone and ProtoNets meta-learner on CUB dataset. Generally, we can see similar trends as we have seen from Fig.4 in the main paper. However, since CUB dataset is fine-grained, the forgetting in \textit{mean accuracy on seen classes} is not as serious as other three coarse-grained datasets in Fig.4. Instead, we can observe the increasing \textit{mean accuracy on seen classes}, that is because the new tasks may benefit from the accumulated knowledge of the old tasks.

\section{Extended comparison with CL methods}

\begin{figure*}

\begin{minipage}[b]{0.48\linewidth}
\centering
\includegraphics[width=\textwidth]{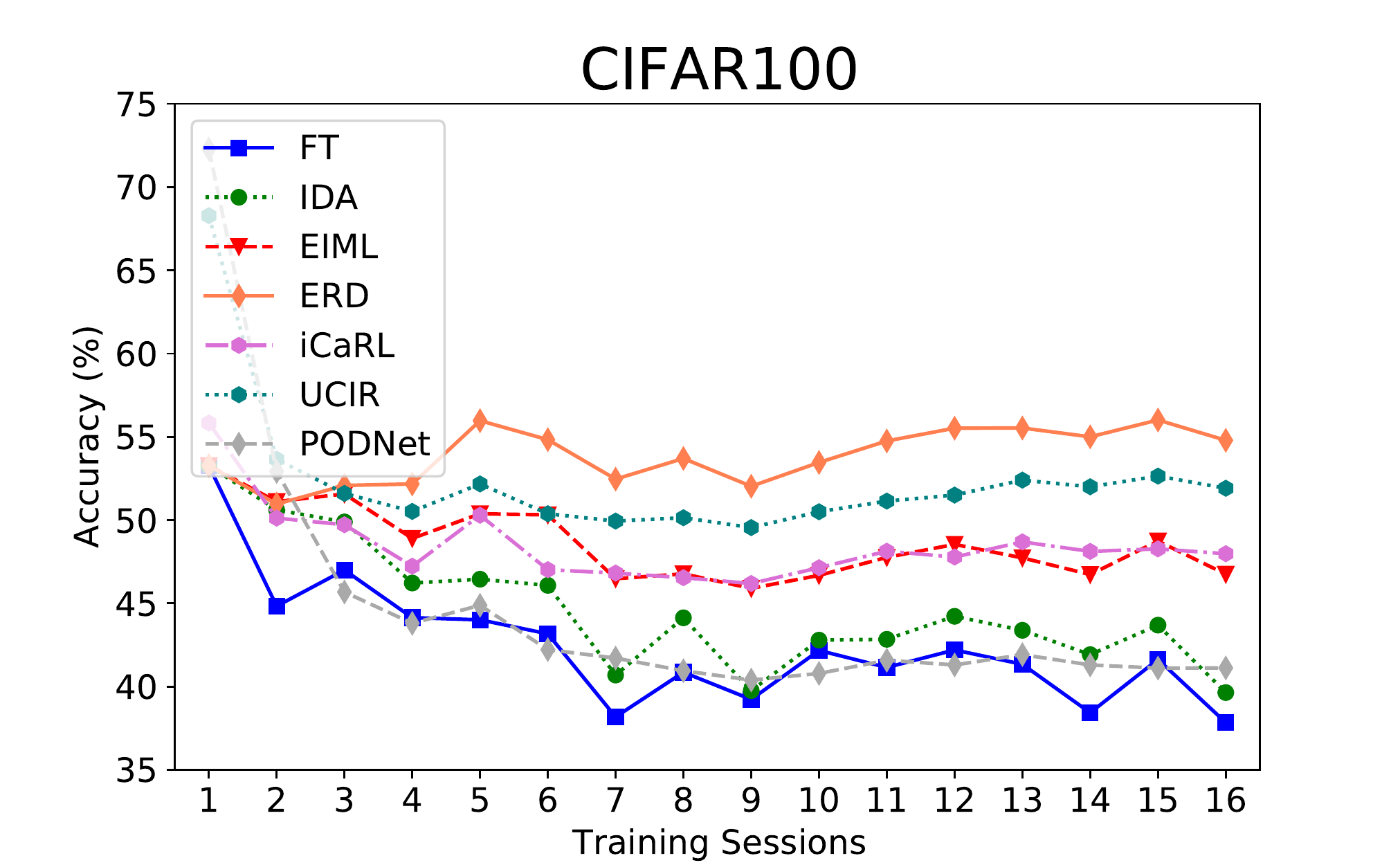}
\subcaption{1-shot mean accuracy on seen classes
}
\label{fig:rebuttal_5shot_16task_cifar100_seen}
\end{minipage}
\hfill%
\begin{minipage}[b]{0.48\linewidth}
\centering
\includegraphics[width=\textwidth]{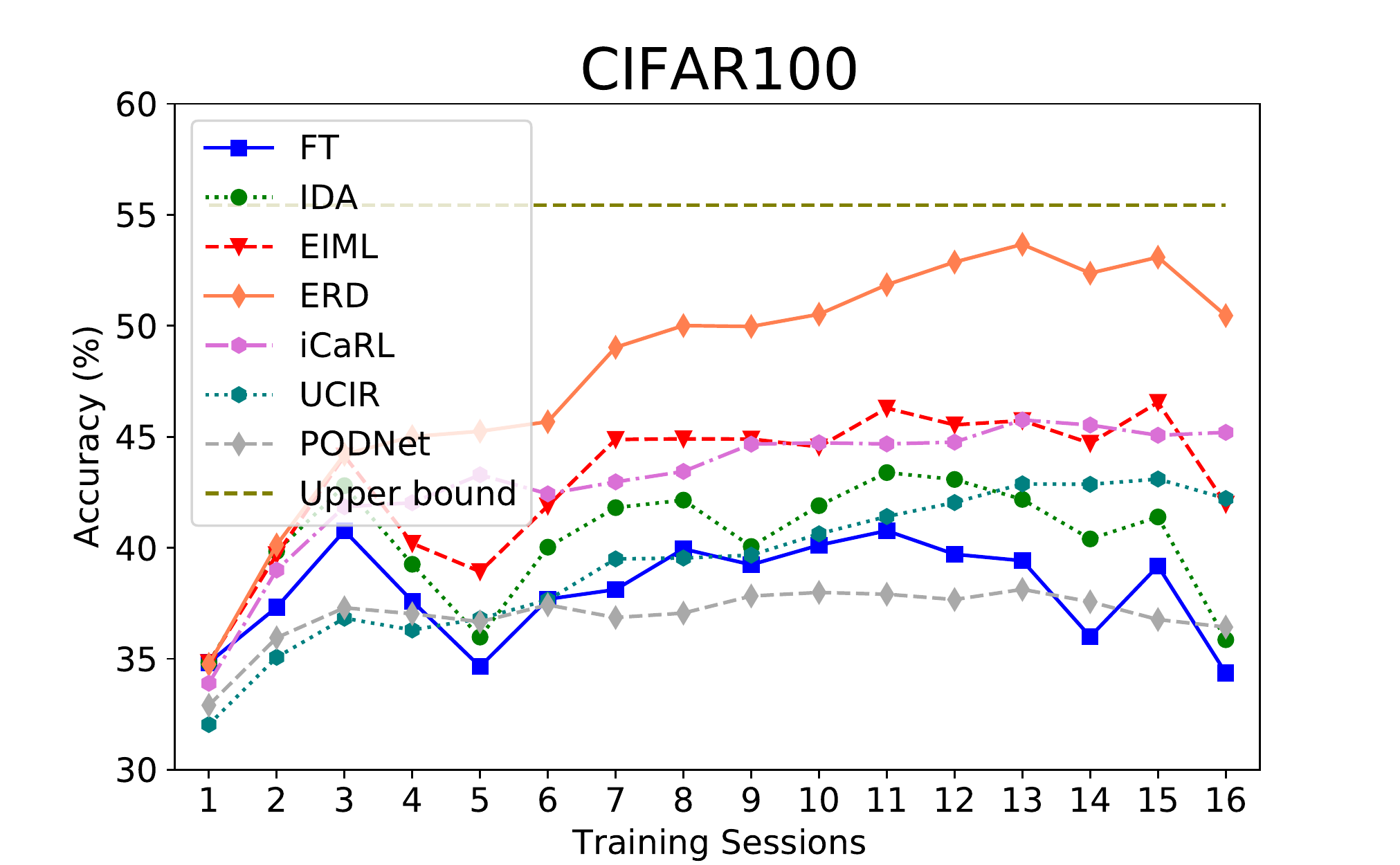}
\subcaption{1-shot meta-test accuracy
}
\label{fig:rebuttal_5shot_16task_cifar100_meta}
\end{minipage}
\begin{minipage}[b]{0.48\linewidth}
\centering
\includegraphics[width=\textwidth]{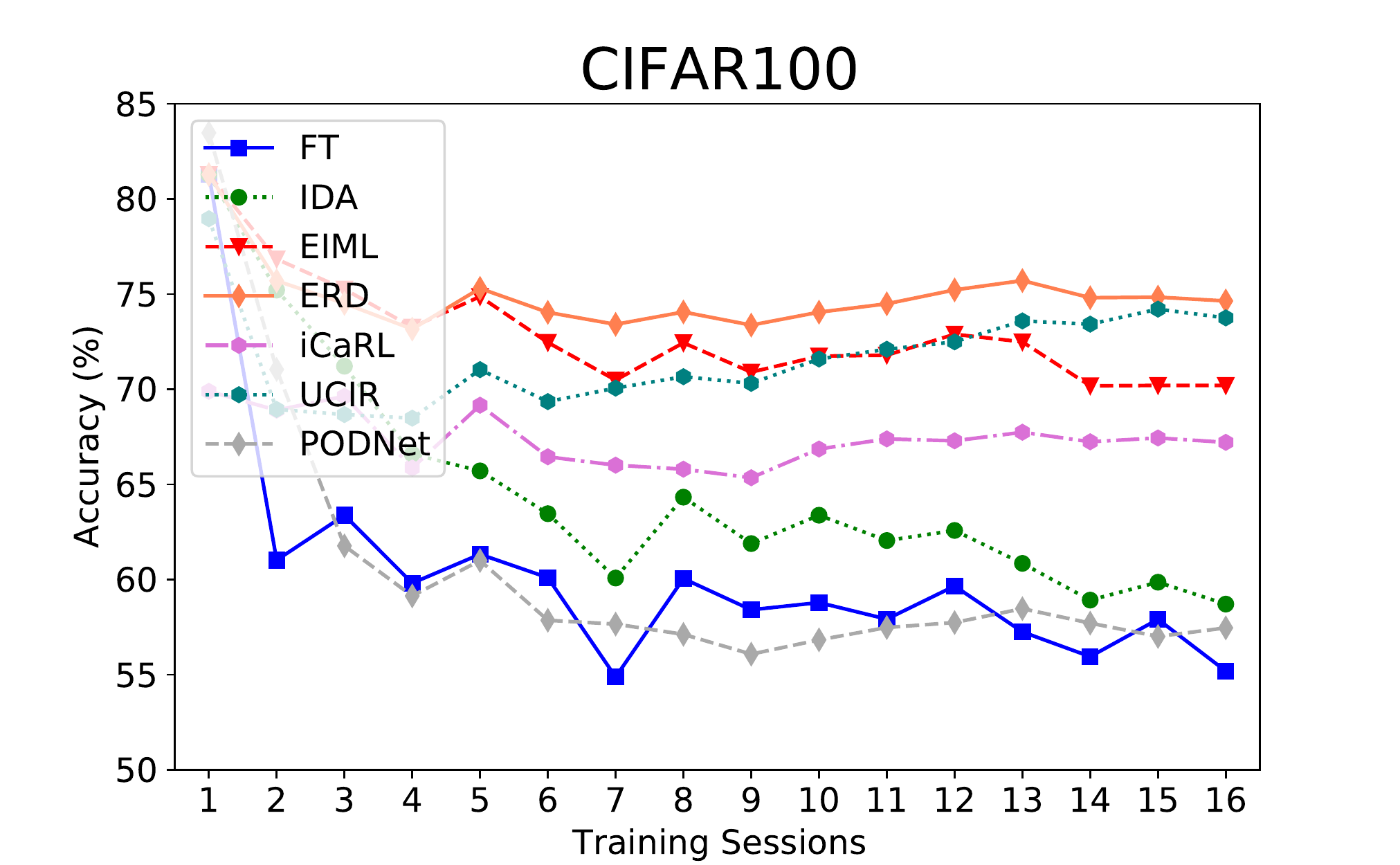}
\subcaption{5-shot mean accuracy on seen classes
}
\label{fig:rebuttal_5shot_16task_cifar100_seen}
\end{minipage}
\hfill%
\begin{minipage}[b]{0.48\linewidth}
\centering
\includegraphics[width=\textwidth]{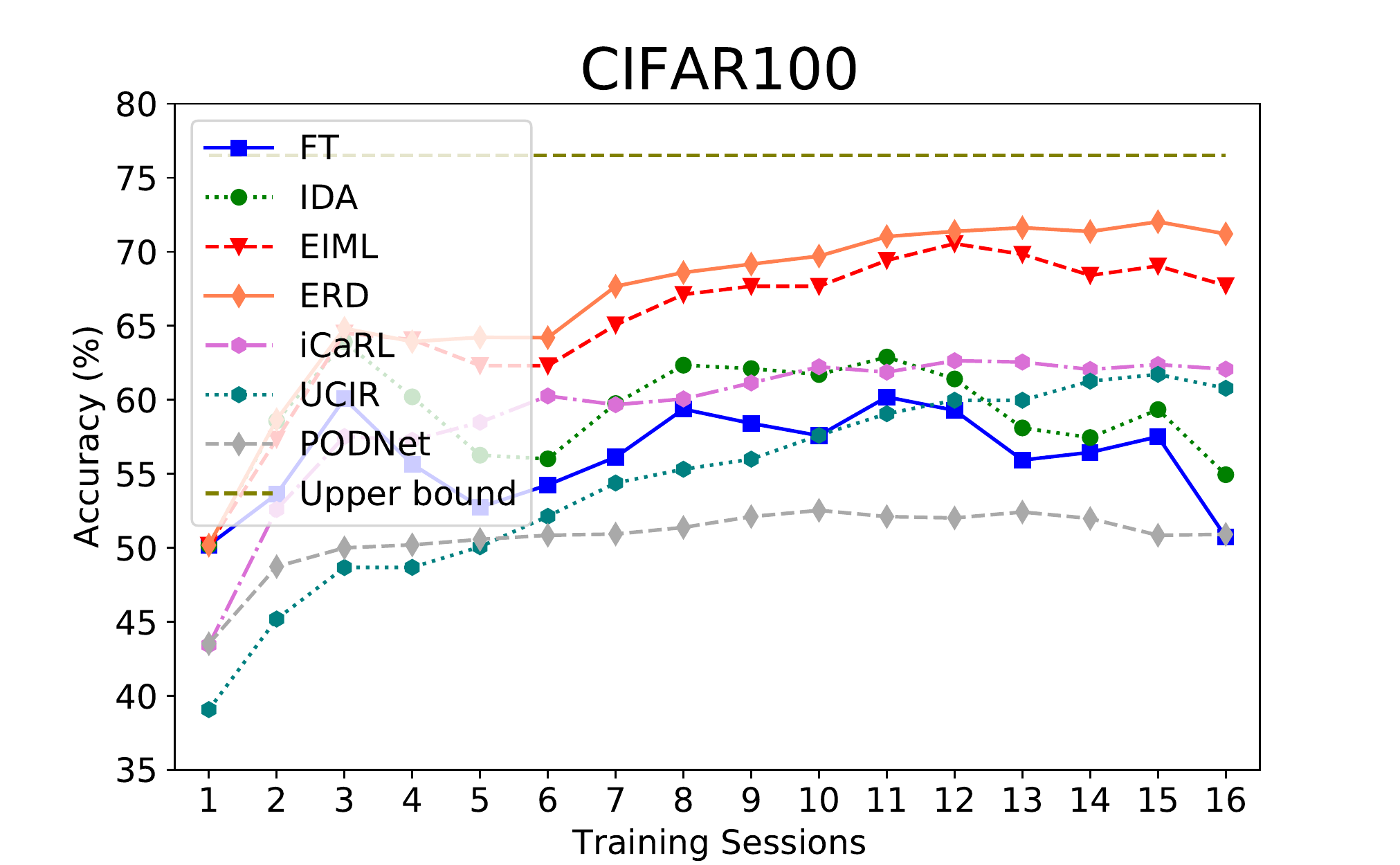}
\subcaption{5-shot meta-test accuracy
}
\label{fig:rebuttal_5shot_16task_cifar100_meta}
\end{minipage}
\caption{
Comparison with CL methods on 1-shot 5-way 16-task setting with a 4-Conv backbone and ProtoNets meta-learner on CIFAR-100. (Left) Mean accuracy on seen classes. (Right) Meta-test accuracy on the unseen meta-test set.
}
\vspace{-0.07in}
\label{fig:rebuttal_16_task_cifar100_CL}
\end{figure*}

Table~\ref{tab:compare_CL_methods} is the full table of Table 3 in the main text. We extend the Table with a comparison of the 5-shot setting, which also shows similar trends as already seen in the 1-shot setting. The conclusion is that for the IML problem regular CL methods are inferior to our proposed ERD. To better visualize the trends, we plot the curves in Fig.~\ref{fig:rebuttal_16_task_cifar100_CL}.

\begin{table}[h]
\begin{center}
\scalebox{0.65}{
\begin{tabular}{|r|ccccc|ccccc|}
\hline
 \multicolumn{11}{|c|}{Datasets: CIFAR100, Learner: ProtoNets, Backbone: 4-Conv} \\

\hline
Evaluation:  & \multicolumn{5}{|c|}{Meta-test accuracy} & \multicolumn{5}{|c|}{Mean accuracy on seen classes} \\ 
\hline\hline
\multicolumn{11}{|c|}{1-shot/5-way \textbf{\textit{16-task}} setting}  \\
\hline
 Sessions:     & 2&4&8&16 & avg  & 2&4&8&16 & avg  \\
\hline\hline
FT     & 37.3 & 37.6 & 40.0 & 34.4 & 38.1 & 44.8 & 44.1 & 40.9 & 37.8 & 42.5  \\ 
IDA    & 39.8 & 39.3 & 42.2 & 35.9 & 40.3 & 50.6 & 46.2 & 44.1 & 39.6 & 44.7  \\ 
EIML   & 39.7 & 40.2 & 44.9 & 42.0 & 43.1 & 51.1 & 48.9 & 46.8 & 46.7 & 48.6  \\ 
ERD    & \textbf{40.1} & \textbf{45.0} & \textbf{50.0} & \textbf{50.5} & \textbf{48.1} & {51.0} & \textbf{52.2} & \textbf{53.7} & \textbf{54.8} & \textbf{53.9}  \\ 
\hline
iCaRL  & 39.0 & 42.0 & 43.4 & 45.2 & 43.1 & 50.1 & 47.2 & 46.5 & 48.0 & 48.5  \\ 
UCIR   & 35.1 & 36.3 & 39.5 & 42.2 & 39.3 & \textbf{53.6} & 50.5 & 50.1 & 51.9 & 52.4  \\ 
PODNet & 36.0 & 37.0 & 37.1 & 36.4 & 37.0 & 52.9 & 43.8 & 41.0 & 41.1 & 44.6  \\
\hline\hline
\multicolumn{11}{|c|}{5-shot/5-way \textbf{\textit{16-task}} setting}  \\
\hline
 Sessions:     & 2&4&8&16 & avg  & 2&4&8&16 & avg  \\

\hline
FT     & 53.6 & 55.7 & 59.4 & 50.7 & 56.1 & 61.0 & 59.8 & 60.1 & 55.2 & 60.2  \\ 

IDA    & 58.6 & 60.2 & 62.3 & 54.9 & 59.1 & 75.2 & 66.6 & 64.3 & 58.7 & 64.8  \\ 

EIML   & 57.3 & \textbf{64.1} & 67.1 & 67.7 & 65.2 & \textbf{76.8} & \textbf{73.3} & 72.4 & 70.2 & 73.0  \\ 

ERD    & \textbf{58.7} & {63.9} & \textbf{68.6} & \textbf{71.2} & \textbf{66.9} & {75.7} & {73.2} & \textbf{74.1} & \textbf{74.6} & \textbf{74.9}  \\ 
\hline
iCaRL  & 52.6 & 57.3 & 60.1 & 62.1 & 59.1 & 68.9 & 65.9 & 65.8 & 67.2 & 67.4  \\ 

UCIR   & 45.2 & 48.7 & 55.3 & 60.8 & 54.4 & 68.9 & 68.5 & 70.7 & 73.7 & 71.7  \\ 

PODNet & 48.7 & 50.2 & 51.4 & 50.9 & 50.7 & 71.0 & 59.1 & 57.1 & 57.5 & 60.5  \\
\hline
\end{tabular}
}
\end{center}
\caption{Meta-test accuracy and mean accuracy as a function of the number of training sessions on the 16-task setting using ProtoNets as the meta learner. We evaluate on CIFAR-100 to compare.}
\label{tab:compare_CL_methods}
\end{table}

\section{Extended experimental results for Table 2}

Due to page limit, we show the 5-shot results under 4-task 5-way setup in Table~\ref{tab:complete_table_3datasets}, which is the full version of Table 2 in the main text. We can observe that under 5-shot setup with 4-Conv or ResNet-12 backbones, our model consistently outperforms other methods.

\begin{table*}
\begin{center}
\scalebox{0.66}{
\begin{tabular}{|r|cccc|cccc|cccc|cccc|cccc|cccc|}
\hline

Learner: & \multicolumn{24}{|c|}{ProtoNets}  \\
\hline

Dataset:& \multicolumn{4}{|c|}{Mini-ImageNet}   & \multicolumn{4}{|c|}{CIFAR100 }  & \multicolumn{4}{|c|}{ CUB} &\multicolumn{4}{|c|}{Mini-ImageNet}  & \multicolumn{4}{|c|}{CIFAR100 }  & \multicolumn{4}{|c|}{ CUB} \\
\hline

Backbone: & \multicolumn{12}{|c|}{4-Conv }  & \multicolumn{12}{|c|}{ResNet-12}  \\
\hline
\multicolumn{25}{|c|}{\textbf{\textit{1-shot 5-way 4-task setting}}}  \\
\hline

& \multicolumn{4}{|c|}{Upper bound: 53.2} & \multicolumn{4}{|c|}{Upper bound: 55.4} & \multicolumn{4}{|c|}{Upper bound: 61.1} & \multicolumn{4}{|c|}{Upper bound: 59.9} & \multicolumn{4}{|c|}{Upper bound: 61.8} & \multicolumn{4}{|c|}{Upper bound: 74.8 }\\
\hline\hline
 Sessions: & 1&2&3&4    & 1&2&3&4   & 1&2&3&4  & 1&2&3&4  & 1&2&3&4&  1&2&3&4\\
\hline\hline

FT &43.8& 	44.1 & 	42.4 & 	37.9 &   44.6 & 45.1 & 48.0 & 45.5  &    45.1 & 54.6 & 54.9 & 58.8     &       45.7 & 45.9 & 42.1 & 37.7   & 47.0 & 45.0 & 51.0 & 44.6 & 53.4 & 64.0 & 63.7 & 66.8\\  
IDA &43.8&  48.3  & 47.2 & 42.3    &     44.6 & 48.0 & 51.3 & 47.6 &   45.1 & \textbf{54.7} & 54.9 & 58.7  &  45.7 & 53.0 & 53.7 & 47.6 & 47.0 & 53.6 & 59.2 & 54.8 & 53.4  & 64.4 & 68.8 & 73.3  \\  
EIML  &43.8 &  48.8  &  49.4  & 47.5 & 44.6 & 48.0 & 52.0 & 51.7 & 45.1 & 53.4 & 55.0& 58.9   &  45.7 & 53.2 & 56.5 & 55.8 & 47.0 & 53.3 & 58.3 & 57.8 & 53.4  & 62.8 & 69.1 & 73.3\\

ERD  &43.8   &\textbf{51.1}  &\textbf{52.3}  &\textbf{53.0}  &  44.6 & \textbf{49.5} & \textbf{53.6} & \textbf{55.1} &  45.1 & 53.9 & \textbf{58.3} & \textbf{60.8}    &    45.7 & \textbf{55.2} & \textbf{58.2} & \textbf{59.3} &  47.0 & \textbf{55.6} & \textbf{61.3} & \textbf{61.4}  & 53.4  & \textbf{66.1} & \textbf{72.4} & \textbf{74.1} \\
\hline

\multicolumn{25}{|c|}{\textbf{\textit{5-shot 5-way 4-task setting}}}  \\
\hline

& \multicolumn{4}{|c|}{Upper bound: 75.1} & \multicolumn{4}{|c|}{Upper bound: 76.5} & \multicolumn{4}{|c|}{Upper bound: 82.5}  & \multicolumn{4}{|c|}{Upper bound: 81.9} & \multicolumn{4}{|c|}{Upper bound: 81.0}  & \multicolumn{4}{|c|}{Upper bound: 91.2} \\
\hline\hline
 Sessions & 1&2&3&4   & 1&2&3&4  & 1&2&3&4 & 1&2&3&4 & 1&2&3&4& 1&2&3&4\\
\hline\hline
FT &63.4 &64.1 &65.2 &62.1     &     67.0 & 68.2 & 71.2 & 67.5  &    69.4 & 73.4 & 74.5 & 76.4   &       66.2 & 65.6 & 62.7 & 61.4 & 69.1 & 68.5 & 70.4 & 66.4 & 76.3 &  80.7 &  80.3 & 84.2 \\  
IDA &63.4 &68.5 &68.1 &66.0   &    67.0 & 70.3 & 72.8 & 69.6  &    69.4 & 75.4 & 76.0 & 78.6  &       66.2 & 73.1 & 75.5 & 74.5 & 69.1& 75.5 & 77.9 & 78.8 & 76.3 & 81.8 & 84.4 & 86.7  \\  
EIML &63.4 &69.1  &70.3 &70.2    &  67.0 & 70.7 & 73.6 & 72.7 &   69.4 & 75.2 & 78.2 & 79.0 &  66.2 & 74.6 & 77.5 & 78.3 & 69.1& 76.8 & 78.6 & 80.3 &76.3 & 83.2& 86.1 & 88.3 \\
ERD  &63.4 &\textbf{69.4} &\textbf{71.4} &\textbf{72.2}&   67.0 & \textbf{71.2} & \textbf{74.4} & \textbf{73.9}  &   69.4 & \textbf{75.9} & \textbf{78.7} & \textbf{80.4}    &     66.2 & \textbf{74.7} & \textbf{77.7} & \textbf{80.0}  & 69.1& \textbf{77.2} & \textbf{79.4} & \textbf{80.8}  &76.3 & \textbf{83.4} & \textbf{86.6} & \textbf{89.6}   \\
\hline

\end{tabular}
}
\end{center}
\caption{Meta-test accuracy by training session in the 4-task setting. We evaluate 1-shot/5-shot 5-way few-shot recognition on Mini-ImageNet, CIFAR-100, and CUB using two different backbones. 
}
\label{tab:complete_table_3datasets}
\end{table*}

